\documentclass[journal]{IEEEtran}
\usepackage{amsthm}
\usepackage{amsmath}
\usepackage{amssymb}
\usepackage{graphicx}
\usepackage{algorithm}
\usepackage{algpseudocode}

\newtheorem{proposition}{Proposition}
\newtheorem{definition}{Definition}

\newcommand{\x}{\boldsymbol{x}}

\newcommand{\R}{\boldsymbol{R}}

\begin{document}
\title{Explaining NonLinear Classification Decisions with Deep Taylor Decomposition}

\author{Gr\'egoire Montavon$^*$,
        Sebastian Bach, 
        Alexander Binder,
        Wojciech Samek$^*$,
        and~Klaus-Robert M\"{u}ller$^*$
\thanks{This work was supported by the Brain Korea 21 Plus Program through the National Research Foundation of Korea funded by the Ministry of Education. This work was also supported by the grant DFG (MU~987/17-1) and by the German Ministry for Education and Research as Berlin Big Data Center BBDC (01IS14013A). This publication only reflects the authors views. Funding agencies are not liable for any use that may be made of the information contained herein. {\it Asterisks indicate corresponding author}.}
\thanks{$^*$G. Montavon is with the Berlin Institute of Technology (TU Berlin), 10587 Berlin, Germany. (e-mail: gregoire.montavon@tu-berlin.de)}
\thanks{S. Bach is with Fraunhofer Heinrich Hertz Institute, 10587 Berlin, Germany. (e-mail: sebastian.bach@hhi.fraunhofer.de)}
\thanks{A. Binder is with the Singapore University of Technology and Design, 487372, Singapore. (e-mail: alexander\_binder@sutd.edu.sg)}
\thanks{$^*$W. Samek is with Fraunhofer Heinrich Hertz Institute, 10587 Berlin, Germany. (e-mail: wojciech.samek@hhi.fraunhofer.de)}
\thanks{$^*$K.-R. M\"uller is with the Berlin Institute of Technology (TU Berlin), 10587 Berlin, Germany, and also with the Department of Brain and Cognitive Engineering, Korea University, Seoul 136-713, Korea (e-mail: klaus-robert.mueller@tu-berlin.de)}}

\maketitle

\begin{abstract}
Nonlinear methods such as Deep Neural Networks (DNNs) are the gold standard for various challenging machine learning problems, e.g., image classification, natural language processing or human action recognition. Although these methods perform impressively well, they have a significant disadvantage, the lack of transparency, limiting the interpretability of the solution and thus the scope of application in practice. Especially DNNs act as black boxes due to their multilayer nonlinear structure. In this paper we introduce a novel methodology for interpreting generic multilayer neural networks by decomposing the network classification decision into contributions of its input elements. Although our focus is on image classification, the method is applicable to a broad set of input data, learning tasks and network architectures. Our method is based on {\em deep} Taylor decomposition and efficiently utilizes the structure of the network by backpropagating the explanations from the output to the input layer. We evaluate the proposed method empirically on the MNIST and ILSVRC data sets.
\end{abstract}

\section{Introduction}

Nonlinear models have been used since the advent of machine learning (ML) methods and are integral part of many popular algorithms. They include, for example, graphical models \cite{jordan1998:learning_graph_mod}, kernels \cite{schoelkopf:book,DBLP:journals/tnn/MullerMRTS01}, Gaussian processes \cite{book:gp}, neural networks \cite{bishop95,nntricks,DBLP:series/lncs/LeCunBOM12}, boosting \cite{book:boosting}, or random forests \cite{DBLP:journals/ml/Breiman01}. Recently, a particular class of nonlinear methods, Deep Neural Networks (DNNs), revolutionized the field of automated image classification by demonstrating impressive performance on large benchmark data sets \cite{DBLP:conf/nips/KrizhevskySH12, DBLP:conf/nips/CiresanGGS12, DBLP:journals/corr/SzegedyLJSRAEVR14}. Deep networks have also been applied successfully to other research fields such as natural language processing \cite{DBLP:journals/jmlr/CollobertWBKKK11,Socher-etal:2013}, human action recognition \cite{DBLP:conf/icml/JiXYY10,DBLP:conf/cvpr/LeZYN11}, or physics \cite{montavon-njp13, baldi14}. Although these models are highly successful in terms of performance, they have a drawback of acting like a {\em black box} in the sense that it is not clear {\em how} and {\em why} they arrive at a particular classification decision. This lack of transparency is a serious disadvantage as it prevents a human expert from being able to verify, interpret, and understand the reasoning of the system.

\begin{figure}
\centering
\includegraphics[width=1.0\linewidth]{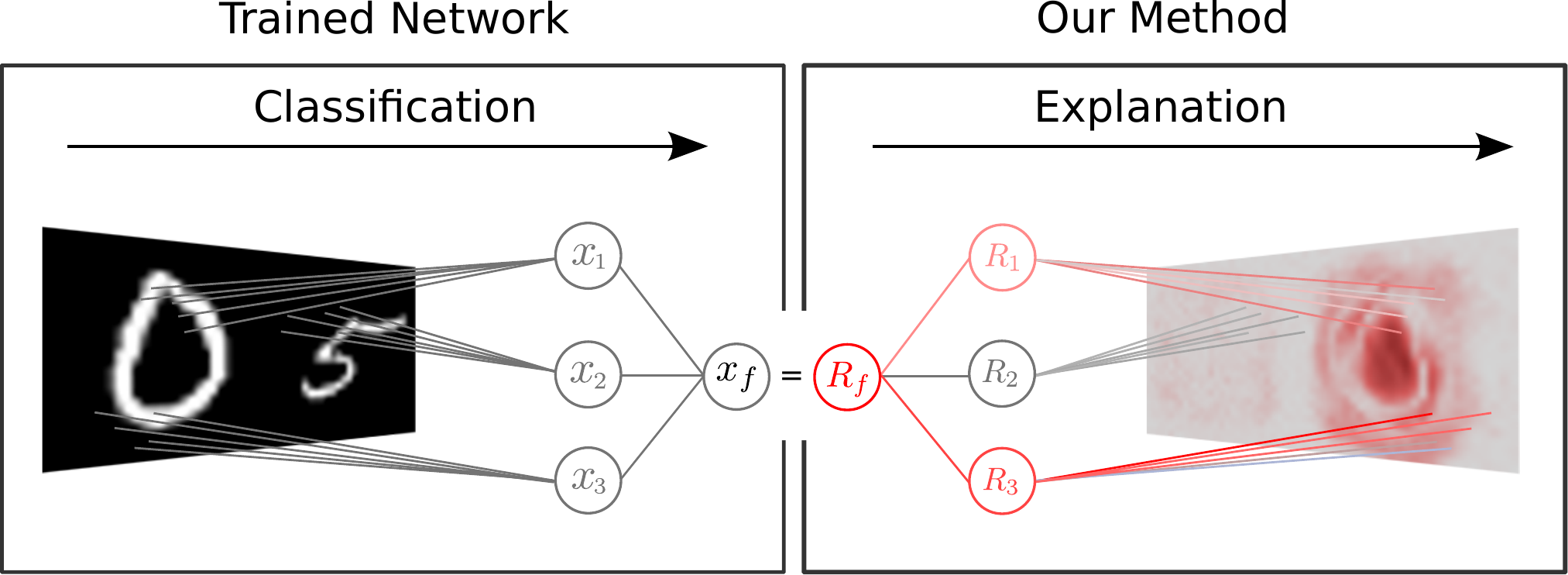}\vskip -2mm
\caption{Overview of our method for explaining a nonlinear classification decision. The method produces a pixel-wise heatmap explaining {\em why} a neural network classifier has come up with a particular decision (here, detecting the digit ``0'' in an input image composed of two digits). The heatmap is the result of a {\em deep} Taylor decomposition of the neural network function. Note that for the purpose of the visualization, the left and right side of the figure are mirrored.}
\label{figure:overview}
\end{figure}

An {\em interpretable classifier} explains its nonlinear classification decision in terms of the inputs. For instance, in image classification problems, the classifier should not only indicate whether an image of interest belongs to a certain category or not, but also explain what structures (e.g. pixels in the image) were the basis for its decision (cf. Figure \ref{figure:overview}). This additional information helps to better assess the quality of a particular prediction, or to verify the overall reasoning ability of the trained classifier. Also, information about which pixels are relevant in a particular image, could be used for determining which region of the image should be the object of further analysis. Linear models readily provide explanations in terms of input variables (see for example \cite{DBLP:journals/neuroimage/HaufeMGDHBB14,Oaxaca1973}). However, because of the limited expressive power of these models, they perform poorly on complex tasks such as image recognition. Extending linear analysis techniques to more realistic nonlinear models such as deep neural networks, is therefore of high practical relevance.

Recently, a significant amount of work has been dedicated to make the deep neural network more transparent to its user, in particular, improving the overall interpretability of the learned model, or explaining individual predictions. For example, Zeiler et al. \cite{DBLP:journals/corr/ZeilerF13} have proposed a network propagation technique to identify patterns in the input data that are linked to a particular neuron activation or a classification decision. Subsequently, Bach et al. \cite{bach15} have introduced the concept of pixel-wise decomposition of a classification decision, and how such decomposition can be achieved either by Taylor decomposition, or by a relevance propagation algorithm. Specifically, the authors distinguish between (1) functional approaches that view the neural network as a {\em function} and disregard its topology, and (2) message passing approaches, where the decomposition stems from a simple {\em propagation rule} applied uniformly to all neurons of the deep network.

The main goal of this paper is to reconcile the functional and rule-based approaches for obtaining these decompositions, in a similar way to the error backpropagation algorithm \cite{rumelhart86} that also has a functional and a message passing interpretation. We call the resulting framework {\em deep Taylor decomposition}. This new technique seeks to replace the analytically intractable standard Taylor decomposition problem by a multitude of simpler analytically tractable Taylor decompositions---one per neuron. The proposed method results in a relevance redistribution process like the one illustrated in Figure \ref{figure:overview} for a neural network trained to detect the digit ``0'' in an image, in presence of another distracting digit. The classification decision is first decomposed in terms of contributions $R_1,R_2,R_3$ of respective hidden neurons $x_1,x_2,x_3$, and then, the contribution of each hidden neuron is independently redistributed onto the pixels, leading to a relevance map (or heatmap) in the pixel space, that explains the classification ``0''.

A main result of this work is the observation that application of deep Taylor decomposition to neural networks used for image classification, yields rules that are similar to those proposed by \cite{bach15} (the $\alpha\beta$-rule and the $\epsilon$-rule), but with specific instantiations of their hyperparameters, previously set heuristically. Because of the theoretical focus of this paper, we do not perform a broader empirical comparison with other recently proposed methods such as \cite{DBLP:journals/corr/SimonyanVZ13} or \cite{DBLP:journals/corr/ZeilerF13}. However, we refer to \cite{samek-arxiv15} for such a comparison.

The paper is organized as follows: Section \ref{section:decomposition} introduces the general idea of decomposition of a classification score in terms of input variables, and how this decomposition arises from Taylor decomposition or deep Taylor decomposition of a classification function. Section \ref{section:onelayer} applies the proposed deep Taylor decomposition method to a simple detection-pooling neural network. Section \ref{section:twolayers} extends the method to deeper networks, by introducing the concept of relevance model and describing how it can be applied to large GPU-trained neural networks without retraining. Several experiments on MNIST and ILSVRC data are provided to illustrate the methods described here. Section \ref{section:conclusion} concludes.

\subsection*{Related Work}

There has been a significant body of work focusing on the analysis and understanding of nonlinear classifiers such as kernel machines \cite{DBLP:journals/jmlr/BraunBM08, DBLP:journals/jmlr/BaehrensSHKHM10, hansen2011visual, DBLP:journals/spm/MontavonBKM13}, neural networks \cite{Garson1991, DBLP:journals/aei/Goh95, DBLP:conf/nips/GoodfellowLSLN09, DBLP:journals/jmlr/MontavonBM11}, or a broader class of nonlinear models \cite{BazenJoutard2013,bach15}. In particular, some recent analyses have focused on the understanding of state-of-the-art GPU-trained convolutional neural networks for image classification \cite{DBLP:journals/corr/SzegedyZSBEGF13, DBLP:journals/corr/SimonyanVZ13,DBLP:journals/corr/ZeilerF13}, offering new insights on these highly complex models.

Some methods seek to provide a general understanding of the trained model, by measuring important characteristics of it, such as the noise and relevant dimensionality of its feature space(s) \cite{DBLP:journals/jmlr/BraunBM08,DBLP:journals/spm/MontavonBKM13,
DBLP:journals/jmlr/MontavonBM11}, its invariance to certain transformations of the data \cite{DBLP:conf/nips/GoodfellowLSLN09} or the role of particular neurons \cite{understanding_techreport}. In this paper, we focus instead on the interpretation of the prediction of {\em individual} data points, for which portions of the trained model may either be relevant or not relevant.

Technically, the methods proposed in \cite{DBLP:journals/jmlr/BaehrensSHKHM10, hansen2011visual} do not explain the decision of a classifier but rather perform sensitivity analysis by computing the gradient of the decision function. This results in an analysis of variations of that function, without however seeking to provide a full explanation why a certain data point has been predicted in a certain way. Specifically, the gradient of a function does not contain information on the saliency of a feature in the data to which the function is applied. Simonyan et al. \cite{DBLP:journals/corr/SimonyanVZ13} incorporate saliency information by multiplying the gradient by the actual data point.

The method proposed by Zeiler and Fergus \cite{DBLP:journals/corr/ZeilerF13} was designed to visualize and understand the features of a convolutional neural network with max-pooling and rectified linear units. The method performs a backpropagation pass on the network, where a set of rules is applied uniformly to all layers of the network, resulting in an assignment of values onto pixels. The method however does not aim to attribute a defined meaning to the assigned pixel values, except for the fact that they should form a visually interpretable pattern. \cite{bach15} proposed a layer-wise propagation method where the backpropagated signal is interpreted as relevance, and obeys a conservation property. The proposed propagation rules were designed according to this property, and were shown quantitatively to better support the classification decision \cite{samek-arxiv15}. However, the practical choice of propagation rules among all possible ones was mainly heuristic and lacked a strong theoretical justification.

A theoretical foundation to the problem of relevance assignment for a classification decision, can be found in the Taylor decomposition of a nonlinear function. The approach was described by Bazen and Joutard \cite{BazenJoutard2013} as a nonlinear generalization of the Oaxaca method in econometrics \cite{Oaxaca1973}. The idea was subsequently introduced in the context of image analysis \cite{bach15,DBLP:journals/corr/SimonyanVZ13} for the purpose of explaining machine learning classifiers. Our paper extends the standard Taylor decomposition in a way that takes advantage of the deep structure of neural networks, and connects it to rule-based propagation methods, such as \cite{bach15}.

As an alternative to propagation methods, spatial response maps \cite{DBLP:journals/corr/FangGISDDGHMPZZ14} build heatmaps by looking at the neural network output while sliding the neural network in the pixel space. Attention models based on neural networks can be trained to provide dynamic relevance assignment, for example, for the purpose of classifying an image from only a few glimpses of it \cite{DBLP:conf/nips/LarochelleH10}. They can also visualize what part of an image is relevant at a given time in some temporal context \cite{DBLP:conf/icml/XuBKCCSZB15}. However, they usually require specific models that are significantly more complex to design and train.

\section{Pixel-Wise Decomposition of a Function}
\label{section:decomposition}

In this section, we will describe the general concept of explaining a neural network decision by redistributing the function value (i.e. neural network output) onto the input variables in an amount that matches the respective contributions of these input variables to the function value. After enumerating a certain number of desirable properties for the input-wise relevance decomposition, we will explain in a second step how the Taylor decomposition technique, and its extension, deep Taylor decomposition, can be applied to this problem. For the sake of interpretability---and because all our subsequent empirical evaluations focus on the problem of image recognition,---we will call the input variables ``pixels'', and use the letter $p$ for indexing them. Also, we will employ the term ``heatmap'' to designate the set of redistributed relevances onto pixels. However, despite the image-related terminology, the method is applicable more broadly to other input domains such as abstract vector spaces, time series, or more generally any type of input domain whose elements can be processed by a neural network.

Let us consider a positive-valued function $f: \mathbb{R}^d \to \mathbb{R}^+$. In the context of image classification, the input $\x \in \mathbb{R}^d$ of this function can be an image. The image can be decomposed as a set of pixel values $\x = \{x_p\}$ where $p$ denotes a particular pixel. The function $f(\x)$ quantifies the presence (or amount) of a certain type of object(s) in the image. This quantity can be for example a probability, or the number of occurrences of the object. A function value $f(\x) = 0$ indicates the absence of such object(s) in the image. On the other hand, a function value $f(\x) > 0$ expresses the presence of the object(s) with a certain probability or in a certain amount.

We would like to associate to each pixel $p$ in the image a {\em relevance score} $R_p(\x)$, that indicates for an image $\x$ to what extent the pixel $p$ contributes to explaining the classification decision $f(\x)$. The relevance of each pixel can be stored in a heatmap denoted by $\R(\x) = \{R_p(\x)\}$ of same dimensions as the image $\x$. The heatmap can therefore also be visualized as an image. In practice, we would like the heatmapping procedure to satisfy certain properties that we define below.

\begin{definition}
\label{def:conservative} A heatmapping $\R(\x)$ is \underline{conservative} if the sum of assigned relevances in the pixel space corresponds to the total relevance detected by the model, that is
\begin{align*}
\forall \x:~f(\x) = \sum_p R_p(\x).
\end{align*}
\end{definition}

\begin{definition}
\label{def:positive}
A heatmapping $\R(\x)$ is p\!\!\underline{\,\,ositive} if all values forming the heatmap are greater or equal to zero, that is:
\begin{align*}
\forall \x,p:~R_p(\x) \geq 0
\end{align*}
\end{definition}

The first property was proposed by \cite{bach15} and ensures that the total redistributed relevance corresponds to the extent to which the object in the input image is detected by the function $f(\x)$. The second property forces the heatmapping to assume that the model is devoid of contradictory evidence (i.e. no pixels can be in contradiction with the presence or absence of the detected object in the image). These two properties of a heatmap can be combined into the notion of {\em consistency}:
\begin{definition}
\label{def:consistent}
A heatmapping $\R(\x)$ is \underline{consistent} if it is conservative \underline{and} positive. That is, it is consistent if it complies with Definitions \ref{def:conservative} and \ref{def:positive}.
\end{definition}
In particular, a consistent heatmap is forced to satisfy $(f(\x)\!=\!0) \Rightarrow (\R(\x)\!=\!\boldsymbol{0})$. That is, in absence of an object to detect, the relevance is forced to be zero everywhere in the image (i.e. empty heatmap), and not simply to have negative and positive relevance in same amount. We will use Definition \ref{def:consistent} as a formal tool for assessing the correctness of the heatmapping techniques proposed in this paper.

It was noted by \cite{bach15} that there may be multiple heatmapping techniques that satisfy a particular definition. For example, we can consider a heatmapping specification that assigns for all images the relevance uniformly onto the pixel grid:
\begin{align}
\forall p: \quad R_p(\x) = \frac1d \cdot f(\x),
\label{eq:conservation-1}
\end{align}
where $d$ is the number of input dimensions. Alternately, we can consider another heatmapping specification where all relevance is assigned to the first pixel in the image:
\begin{align}
R_p(\x) =
\left\{
\begin{array}{ll}
f(\x) & \quad \text{if}~p= 1\text{st pixel}\\
0     & \quad \text{else}.
\end{array}
\right.
\label{eq:conservation-2}
\end{align}
Both \eqref{eq:conservation-1} and \eqref{eq:conservation-2} are consistent in the sense of Definition \ref{def:consistent}, however they lead to different relevance assignments. In practice, it is not possible to specify explicitly all properties that a heatmapping technique should satisfy in order to be meaningful. Instead, it can be given {\em implicitly} by the choice of a particular algorithm (e.g. derived from a particular mathematical model), subject to the constraint that it complies with the definitions above.

\subsection{Taylor Decomposition}
\label{section:taylor}

\begin{figure}
\centering
\includegraphics[width=1.0\linewidth]{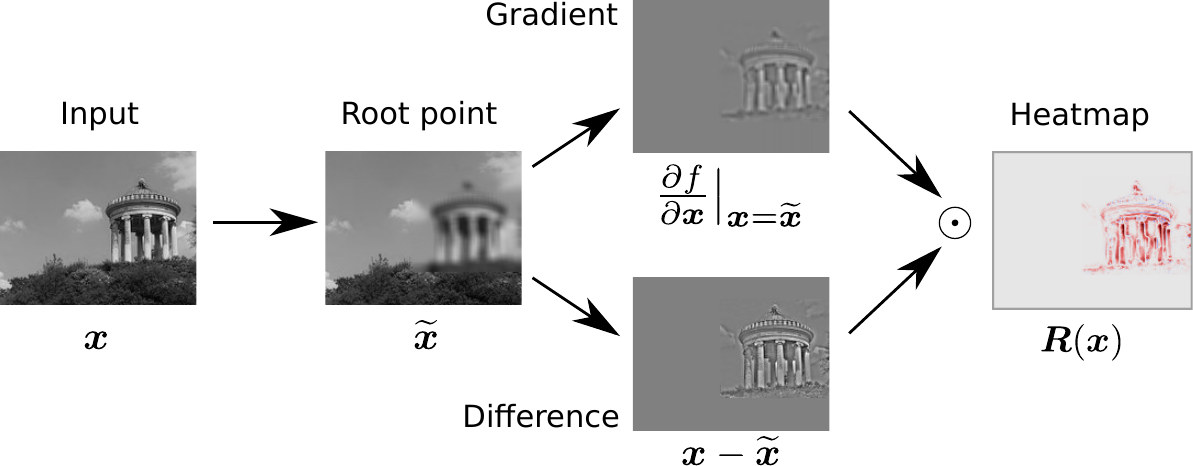}\vskip -2mm
\caption{Cartoon showing the construction of a Taylor-based heatmap from an image $\x$ and a hypothetical function $f$ detecting the presence of objects of class ``building'' in the image. In the heatmap, positive values are shown in red, and negative values are shown in blue.}
\label{fig:heatmap}
\end{figure}

We present a heatmapping method for explaining the classification $f(\x)$ of a data point $\x$, that is based on the Taylor expansion of the function $f$ at some well-chosen {\em root point} $\widetilde \x$, where $f(\widetilde \x) = 0$. The first-order Taylor expansion of the function is given as
\begin{align}
f(\x)
&= f(\widetilde \x) + \left(\frac{\partial f }{\partial \x}\Big|_{\x = \widetilde \x}\right)^{\!\top} \!\! \cdot (\x-\widetilde \x) + \varepsilon \notag\\
&= 0 + \sum_{p} \underbrace{ \frac{\partial f }{\partial x_p}\Big|_{\x = \widetilde \x} \!\!\cdot ( x_p-\widetilde x_p )}_{R_p(\x)} + \, \varepsilon,
\label{eq:taylordecomposition}
\end{align}
where the sum $\sum_p$ runs over all pixels in the image, and $\{ \widetilde x_p \}$ are the pixel values of the root point $\widetilde \x$. We identify the summed elements as the relevances $R_p(\x)$ assigned to pixels in the image. The term $\varepsilon$ denotes second-order and higher-order terms. Most of the terms in the higher-order expansion involve several pixels at the same time and are therefore more difficult to redistribute. Thus, for simplicity, we will consider only the first-order terms for heatmapping. The heatmap (composed of all identified pixel-wise relevances) can be written as the element-wise product ``$\odot$'' between the gradient of the function $\partial f / \partial \x$ at the root point $\widetilde \x$ and the difference between the image and the root $(\x - \widetilde \x)$:
\begin{align*}
\R(\x) = \frac{\partial f }{\partial \x} \Big|_{\x = \widetilde \x} \odot (\x - \widetilde \x).
\end{align*}
Figure~\ref{fig:heatmap} illustrates the construction of a heatmap in a cartoon example, where a hypothetical function $f$ detects the presence of an object of class ``building'' in an image $\x$. In this example, the root point $\widetilde \x$ is the same image as $\x$ where the building has been blurred. The root point $\widetilde \x$ plays the role of a neutral data point that is similar to the actual data point $\x$ but lacks the particular object in the image that causes $f(\x)$ to be positive. The difference between the image and the root point $(\x - \widetilde \x)$ is therefore an image with only the object ``building''. The gradient $\partial f/\partial \x|_{\x = \widetilde \x}$ measures the sensitivity of the class ``building'' to each pixel when the classifier $f$ is evaluated at the root point $\widetilde \x$. Finally, the sensitivities are multiplied element-wise with the difference $(\x - \widetilde \x)$, producing a heatmap that identifies the most contributing pixels for the object ``building''. Strictly speaking, for images with multiple color channels (e.g.\ RGB), the Taylor decomposition will be performed in terms of pixels {\em and} color channels, thus forming multiple heatmaps (one per color channel). Since we are here interested in pixel contributions and not color contributions, we sum the relevance over all color channels, and obtain as a result a single heatmap.

For a given classifier $f(\x)$, the Taylor decomposition approach described above has one free variable: the choice of the root point $\widetilde \x$ at which the Taylor expansion is performed. The example of Figure \ref{fig:heatmap} has provided some intuition on what are the properties of a good root point. In particular, a good root point should selectively remove information from some pixels (here, pixels corresponding to the building at the center of the image), while keeping the surroundings unchanged. This allows in principle for the Taylor decomposition to produce a complete explanation of the detected object which is also insensitive to the surrounding trees and sky.

More formally, a {\em good} root point is one that removes the object (e.g. as detected by the function $f(\x)$, but that minimally deviates from the original point $\x$. In mathematical terms, it is a point $\widetilde \x$ with $f(\widetilde \x)=0$ that lies in the vicinity of $\x$ under some distance metric, for example the nearest root. If $\x,\widetilde \x \in \mathbb{R}^d$, one can show that for a continuously differentiable function $f$ the gradient at the nearest root always points to the same direction as the difference $\x - \widetilde \x$, and their element-wise product is always positive, thus satisfying Definition \ref{def:positive}. Relevance conservation in the sense of Definition \ref{def:conservative} is however not satisfied for general functions $f$ due to the possible presence of non-zero higher-order terms in $\varepsilon$. The nearest root $\widetilde \x$ can be obtained as a solution of an optimization problem \cite{DBLP:journals/corr/SzegedyZSBEGF13}, by minimizing the objective
\begin{align*}
\min_{\boldsymbol{\xi}}~\|\boldsymbol{\xi} - \x\|^2
\quad \text{subject to} \quad
f(\boldsymbol{\xi}) = 0 \quad \text{and} \quad \boldsymbol{\xi} \in \mathcal{X},
\end{align*}
where $\mathcal{X}$ is the input domain. The nearest root $\widetilde \x$ must therefore be obtained in the general case by an iterative minimization procedure. It is time consuming when the function $f(\x)$ is expensive to evaluate or differentiate. Furthermore, it is not necessarily solvable due to the possible non-convexity of the minimization problem.

We introduce in the next sections two variants of Taylor decomposition that seek to avoid the high computational requirement, and to produce better heatmaps. The first one called sensitivity analysis makes use of a single gradient evaluation of the function at the data point. The second one called deep Taylor decomposition exploits the structure of the function $f(\x)$ when the latter is a deep neural network in order to redistribute relevance onto pixels using a single forward-backward pass on the network.

\subsection{Sensitivity Analysis}
\label{section:sensitivity}

A simple method to assign relevance onto pixels is to set it proportional to the squared derivatives of the classifier \cite{Gevrey2003249}:
$$
R(\x) \propto \Big(\frac{\partial f}{\partial \x}\Big)^2,
$$
where the power applies element-wise. This redistribution can be viewed as a special instance of Taylor decomposition where one expands the function at a point $\boldsymbol{\xi} \in \mathbb{R}^d$, which is taken at an infinitesimally small distance from the actual point $\x$, in the direction of maximum descent of $f$ (i.e. $\boldsymbol{\xi} = \x - \delta \cdot \partial f / \partial \x$ with $\delta$ small). Assuming that the function is locally linear, and therefore, the gradient is locally constant, we get
\begin{align*}
f(\x)
&= f(\boldsymbol{\xi}) + \Big(\frac{\partial f }{\partial \x} \Big|_{\x=\boldsymbol{\xi}}\Big)^{\!\top} \!\! \cdot \Big(\x- \Big(\x - \delta \frac{\partial f}{\partial \x}\Big)\Big) + 0\\
&= f(\boldsymbol{\xi}) + \delta \Big(\frac{\partial f }{\partial \x}\Big)^{\!\top} \frac{\partial f }{\partial \x} + 0\\
&= f(\boldsymbol{\xi}) + \sum_p \underbrace{\delta \Big(\frac{\partial f }{\partial x_p}\Big)^{\!2}}_{R_p} + 0,
\end{align*}
where the second-order terms are zero because of the local linearity. The resulting heatmap is positive, but not conservative since almost all relevance is absorbed by the zero-order term $f(\boldsymbol{\xi})$, which is not redistributed. Sensitivity analysis only measures a {\em local} effect and does provide a full explanation of a classification decision. In that case, only relative contributions between different values of $R_p$ are meaningful.

\begin{figure*}
\centering
\includegraphics[width=0.98\linewidth]{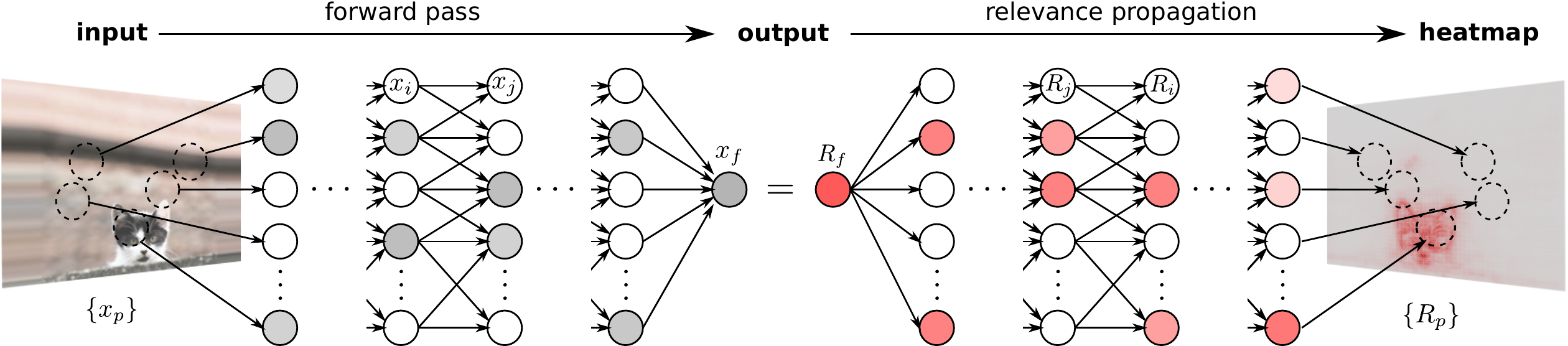}
\caption{Graphical depiction of the computational flow of deep Taylor decomposition. A score $f(\x)$ indicating the presence of the class ``cat'' is obtained by forward-propagation of the pixel values $\{x_p\}$ into a neural network. The function value is encoded by the output neuron $x_f$. The output neuron is assigned relevance $R_f = x_f$. Relevances are backpropagated from the top layer down to the input, where $\{R_p\}$ denotes the relevance scores of all pixels. The last neuron of the lowest hidden layer is perceived as relevant by higher layers and redistributes its assigned relevance onto the pixels. Other neurons of the same layer are perceived as less relevant and do not significantly contribute to the heatmap.
}
\label{figure:neuralnet}
\end{figure*}

\subsection{Deep Taylor Decomposition}
\label{section:deeptaylor}

A rich class of functions $f(\x)$ that can be trained to map input data to classes is the deep neural network (DNN). A deep neural network is composed of multiple layers of representation, where each layer is composed of a set of neurons. The neural network is trained by adapting its set of parameters at each layer, so that the overall prediction error is minimized. As a result of training a deep network, a particular structure or factorization of the learned function emerges \cite{DBLP:conf/icml/LeeGRN09}. For example, each neuron in the first layer may react to a particular pixel activation pattern that is localized in the pixel space. The resulting neuron activations may then be used in higher layers to compose more complex nonlinearities \cite{DBLP:journals/ftml/Bengio09} that involve a larger number of pixels.

The deep Taylor decomposition method presented here is inspired by the divide-and-conquer paradigm, and exploits the property that the function learned by a deep network is structurally decomposed into a set of simpler subfunctions that relate quantities in adjacent layers. Instead of considering the whole neural network function $f$, we consider the mapping of a set of neurons $\{x_i\}$ at a given layer to the relevance $R_j$ assigned to a neuron $x_j$ in the next layer. Assuming that these two objects are functionally related by some function $R_j(\{x_i\})$, we would like to apply Taylor decomposition on this local function in order to redistribute relevance $R_j$ onto lower-layer relevances $\{R_i\}$. For these simpler subfunctions, Taylor decomposition should be made easier, in particular, root points should be easier to find. Running this redistribution procedure in a backward pass leads eventually to the pixel-wise relevances $\{R_p\}$ that form the heatmap.

Figure \ref{figure:neuralnet} illustrates in details the procedure of layer-wise relevance propagation on a cartoon example where an image of a cat is presented to a hypothetical deep network. If the neural network has been trained to detect images with an object ``cat'', the hidden layers have likely implemented a factorization of the pixels space, where neurons are modeling various features at various locations. In such factored network, relevance redistribution is easier in the top layer where it has to be decided which neurons, and not pixels, are representing the object ``cat''. It is also easier in the lower layer where the relevance has already been redistributed by the higher layers to the neurons corresponding to the location of the object ``cat''.

Assuming the existence of a function that maps neuron activities $\{ x_i \}$ to the upper-layer relevance $R_j$, and of a neighboring root point $\{ \widetilde x_i\}$ such that $R_j(\{ \widetilde x_i\}) = 0$, we can then write the Taylor decomposition of $\sum_j R_j$ at $\{x_i\}$ as
\begin{align}
\sum_j R_j
&= \bigg(\frac{\partial \big(\sum_j R_j\big)}{\partial \{x_i\}}\Big|_{\{ \widetilde x_i\}}\bigg)^{\!\top} \!\! \cdot (\{x_i\}-\{ \widetilde x_i\}) + \varepsilon
\nonumber\\
&= \sum_i \underbrace{\sum_j \frac{\partial R_j}{\partial x_i}\Big|_{\{ \widetilde x_i\}} \!\! \cdot (x_i-\widetilde x_i)}_{R_{i}} + \varepsilon,
\label{eq:taylorprop}
\end{align}
that redistributes relevance from one layer to the layer below, where $\varepsilon$ denotes the Taylor residual, where $\big|_{\{ \widetilde x_i\}}$ indicates that the derivative has been evaluated at the root point $\{ \widetilde x_i\}$, where $\sum_j$ runs over neurons at the given layer, and where $\sum_i$ runs over neurons in the lower layer. Equation \ref{eq:taylorprop} allows us to identify the relevance of individual neurons in the lower layer in order to apply the same Taylor decomposition technique one layer below.

If each local Taylor decomposition in the network is conservative in the sense of Definition \ref{def:conservative}, then, the chain of equalities $R_f = \hdots = \sum_j R_j = \sum_i R_i = \hdots = \sum_p R_p$ should hold. This chain of equalities is referred by \cite{bach15} as layer-wise relevance conservation. Similarly, if Definition \ref{def:positive} holds for each local Taylor decomposition, the positivity of relevance scores at each layer $R_f,\dots,\{R_j\},\{R_i\},\dots,\{R_p\} \geq 0$ is also ensured. Finally, if all Taylor decompositions of local subfunctions are consistent in the sense of Definition \ref{def:consistent}, then, the whole deep Taylor decomposition is also consistent in the same sense.

\section{Application to One-Layer Networks}
\label{section:onelayer}

\begin{figure}
\centering
\includegraphics[scale=0.65]{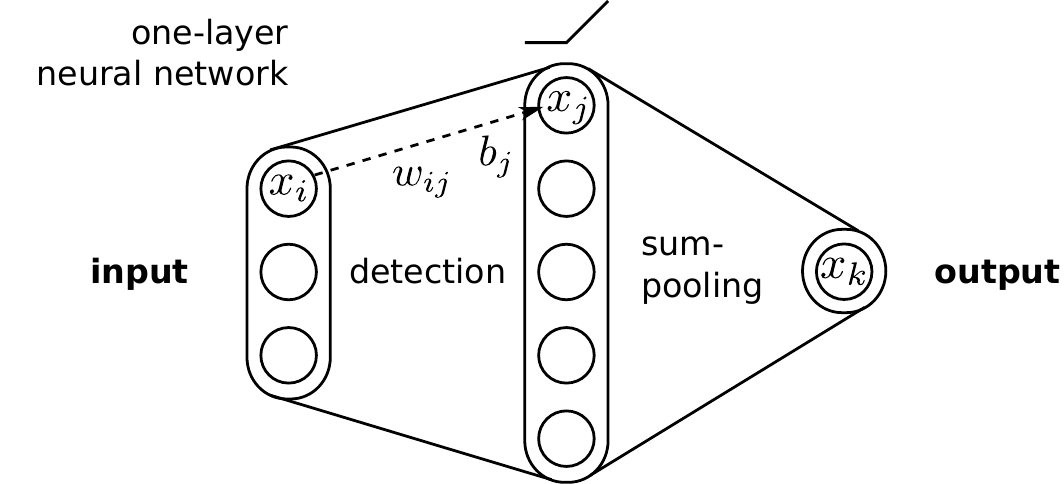}
\vskip -2mm
\caption{\label{fig:onelayer} Detection-pooling network that implements Equations \ref{eq:onelayer-1} and \ref{eq:onelayer-2}: The first layer detects features in the input space, the second layer pools the detected features into an output score.}
\end{figure}

As a starting point for better understanding deep Taylor decomposition, in particular, how it leads to practical rules for relevance propagation, we work through a simple example, with advantageous analytical properties. We consider a detection-pooling network made of one layer of nonlinearity. The network is defined as
\begin{align}
x_j &= \max\big(0,\textstyle{\sum}_i x_i w_{ij} + b_j\big)  \label{eq:onelayer-1}\\
x_k &= \textstyle{\sum_j} x_j  \label{eq:onelayer-2}
\end{align}
where $\{x_i\}$ is a $d$-dimensional input, $\{x_j\}$ is a detection layer, $x_k$ is the output, and $\theta = \{w_{ij},b_j\}$ are the weight and bias parameters of the network. The one-layer network is depicted in Figure \ref{fig:onelayer}. The mapping $\{x_i\} \to x_k$ defines a function $g \in \mathcal{G}$, where $\mathcal{G}$ denotes the set of functions representable by this one-layer network. We will set an additional constraint on biases, where we force $b_j \leq 0$ for all $j$. Imposing this constraint guarantees the existence of a root point $\{\widetilde x_i\}$ of the function $g$ (located at the origin), and thus also ensures the applicability of standard Taylor decomposition, for which a root point is needed.

We now perform the deep Taylor decomposition of this function. We start by equating the predicted output to the amount of total relevance that must be backpropagated. That is, we define $R_k = x_k$. The relevance for the top layer can now be expressed in terms of lower-layer neurons as:
\begin{align}
R_k  &= \textstyle{\sum}_j x_j
\label{eq:pooling-rel}
\end{align}
Having established the mapping between $\{x_j\}$ and $R_k$, we would like to redistribute $R_k$ onto neurons $\{x_j\}$. Using Taylor decomposition (Equation \ref{eq:taylordecomposition}), redistributed relevances $R_j$ can be written as:
\begin{align}
R_j = \frac{\partial R_k}{\partial x_j}\Big|_{\{\widetilde x_j \}} \cdot (x_j - \widetilde x_j).
\label{eq:toplayer-taylor}
\end{align}
We still need to choose a root point $\{\widetilde x_j\}$. The list of all root points of this function is given by the plane equation $\sum_j \widetilde x_j = 0$. However, for the root to play its role of reference point, it should be admissible. Here, because of the application of the function $\max(0,\cdot)$ in the preceding layer, the root point must be positive. The only point that is both a root ($\sum_j \widetilde x_j = 0$) and admissible ($\forall j: \widetilde x_j \geq 0$) is $\{\widetilde x_j\} = \boldsymbol{0}$. Choosing this root point in Equation \ref{eq:toplayer-taylor}, and observing that the derivative $\frac{\partial R_k}{\partial x_j} = 1$, we obtain the first rule for relevance redistribution:
\begin{align}
\boxed{R_j = x_j}
\label{eq:relprop-x}
\end{align}
In other words, the relevance must be redistributed on the neurons of the detection layer in same proportion as their activation value. Trivially, we can also verify that the relevance is conserved during the redistribution process ($\sum_j R_j = \sum_j x_j = R_k$) and positive ($R_j = x_j \geq 0$).

Let us now express the relevance $R_j$ as a function of the input neurons $\{x_i\}$. Because $R_j = x_j$ as a result of applying the propagation rule of Equation \ref{eq:relprop-x}, we can write
\begin{align}
R_j = \max\big(0,\textstyle{\sum}_i x_i w_{ij} + b_j\big),
\label{eq:onelayer-rel}
\end{align}
that establishes a mapping between $\{x_i\}$ and $R_j$. To obtain redistributed relevances $\{R_i\}$, we will apply Taylor decomposition again on this new function. The identification of the redistributed total relevance $\sum_j R_j$ onto the preceding layer was identified in Equation \ref{eq:taylorprop} as:
\begin{align}
R_i &= \sum_j \frac{\partial R_j}{\partial x_i} \Big|_{\{\widetilde x_i\}^{(j)}} \cdot (x_i - \widetilde x_i^{(j)}).
\label{eq:onelayer}
\end{align}
Relevances $\{R_i\}$ can therefore be obtained by performing as many Taylor decompositions as there are neurons in the hidden layer. Note that a superscript $^{(j)}$ has been added to the root point $\{\widetilde x_{i}\}$ in order to emphasize that a different root point is chosen for decomposing each relevance $R_j$. We will introduce below various methods for choosing a root $\{\widetilde x_i\}^{(j)}$ that consider the diversity of possible input domains $\mathcal{X} \subseteq \mathbb{R}^d$ to which the data belongs. Each choice of input domain and associated method to find a root will lead to a different rule for propagating relevance $\{R_j\}$ to $\{R_i\}$.

\subsection{Unconstrained Input Space and the $w^2$-Rule}

We first consider the simplest case where any real-valued input is admissible ($\mathcal{X} = \mathbb{R}^d$). In that case, we can always choose the root point $\{\widetilde x_{i}\}^{(j)}$ that is nearest in the Euclidean sense to the actual data point $\{x_i\}$. When $R_j > 0$, the nearest root of $R_j$ as defined in Equation \ref{eq:onelayer-rel} is the intersection of the plane equation $\sum_i \widetilde x_{i}^{(j)} w_{ij} + b_j = 0$, and the line of maximum descent $\{\widetilde x_i\}^{(j)} = \{x_i\} + t \cdot \boldsymbol{w}_{j}$, where $\boldsymbol{w}_{j}$ is the vector of weight parameters that connects the input to neuron $x_j$ and $t \in \mathbb{R}$. The intersection of these two subspaces is the nearest root point. It is given by $\{\widetilde x_{i}\}^{(j)} = \{x_i - \frac{w_{ij}}{\sum_i w_{ij}^2} (\sum_i x_i w_{ij} + b_j)\}$. Injecting this root into Equation \ref{eq:onelayer}, the relevance redistributed onto neuron $i$ becomes:
\begin{align}
\boxed{R_i = \sum_j \frac{w_{ij}^2}{\sum_{i'} w_{i'j}^2} R_j}
\label{eq:relprop-w2}
\end{align}
The propagation rule consists of redistributing relevance according to the square magnitude of the weights, and pooling relevance across all neurons $j$. This rule is also valid for $R_j = 0$, where the actual point $\{x_i\}$ is already a root, and for which no relevance needs to be propagated.
\begin{proposition}
For all $g \in \mathcal{G}$, the deep Taylor decomposition with the $w^2$-rule is consistent in the sense of Definition \ref{def:consistent}.
\label{prop:w2rule-consistent}
\end{proposition}
The $w^2$-rule resembles the rule by \cite{Garson1991,Gevrey2003249} for determining the importance of input variables in neural networks, where absolute values of $w_{ij}$ are used in place of squared values. It is important to note that the decomposition that we propose here is modulated by the upper layer data-dependent $R_j$s, which leads to an individual explanation for each data point.

\subsection{Constrained Input Space and the $z$-Rules}

When the input domain is restricted to a subset $\mathcal{X} \subset \mathbb{R}^d$, the nearest root of $R_j$ in the Euclidean sense might fall outside of $\mathcal{X}$. In the general case, finding the nearest root in this constrained input space can be difficult. An alternative is to further restrict the search domain to a subset of $\mathcal{X}$ where nearest root search becomes feasible again.

We first study the case $\mathcal{X} = \mathbb{R}_+^d$, which arises, for example in feature spaces that follow the application of rectified linear units. In that case, we restrict the search domain to the segment $(\{x_i 1_{w_{ij}<0}\},\{x_i\}) \subset \mathbb{R}_+^d$, that we know contains at least one root at its first extremity. Injecting the nearest root on that segment into Equation \ref{eq:onelayer}, we obtain the relevance propagation rule:
\begin{align*}
\boxed{R_{i} = \sum_j \frac{z_{ij}^+}{\sum_{i'} z_{i'j}^+} R_{j}}
\end{align*}
(called $z^+\!$-rule), where $z_{ij}^+ = x_i w_{ij}^+$, and where $w_{ij}^+$ denotes the positive part of $w_{ij}$. This rule corresponds for positive input spaces to the $\alpha\beta$-rule formerly proposed by \cite{bach15} with $\alpha=1$ and $\beta=0$. The $z^+\!$-rule will be used in Section \ref{section:twolayers} to propagate relevances in higher layers of a neural network where neuron activations are positive.

\begin{proposition}
For all $g \in \mathcal{G}$ and data points $\{x_i\} \in \mathbb{R}_+^d$, the deep Taylor decomposition with the $z^+\!$-rule is consistent in the sense of Definition \ref{def:consistent}.
\label{prop:zprule-consistent}
\end{proposition}

For image classification tasks, pixel spaces are typically subjects to box-constraints, where an image has to be in the domain $\mathcal{B} = \{ \{x_i\} : \forall_{i=1}^d~l_i \leq x_i \leq h_i \}$, where $l_i \leq 0$ and $h_i \geq 0$ are the smallest and largest admissible pixel values for each dimension. In that new constrained setting, we can restrict the search for a root on the segment $(\{l_i 1_{w_{ij}>0} + h_i 1_{w_{ij}<0}\},\{x_i\}) \subset \mathcal{B}$, where we know that there is at least one root at its first extremity. Injecting the nearest root on that segment into Equation \ref{eq:onelayer}, we obtain relevance propagation rule:
\begin{align*}
\boxed{R_{i} = \sum_j \frac{z_{ij} - l_i w_{ij}^+ - h_i w_{ij}^-}{\sum_{i'} z_{i'j} - l_i w_{i'j}^+ - h_i w_{i'j}^-} R_{j}}
\end{align*}
(called $z^\mathcal{B}\!$-rule), where $z_{ij} = x_i w_{ij}$, and where we note the presence of data-independent additive terms in the numerator and denominator. The idea of using an additive term in the denominator was formerly proposed by \cite{bach15} and called $\epsilon$-stabilized rule. However, the objective of \cite{bach15} was to make the denominator non-zero to avoid numerical instability, while in our case, the additive terms serve to enforce positivity.

\begin{proposition}
For all $g \in \mathcal{G}$ and data points $\{x_i\} \in \mathcal{B}$, the deep Taylor decomposition with the $z^\mathcal{B}\!$-rule is consistent in the sense of Definition \ref{def:consistent}.
\label{prop:zbrule-consistent}
\end{proposition}

Detailed derivations of the proposed rules, proofs of Propositions \ref{prop:w2rule-consistent}, \ref{prop:zprule-consistent} and \ref{prop:zbrule-consistent}, and algorithms that implement these rules efficiently are given in the supplement. The properties of the relevance propagation techniques considered so far (when applied to functions $g \in \mathcal{G}$), their domain of applicability, their consistency, and other computational properties, are summarized in the table below:

\begin{center}
\begin{tabular}{l||c|c|c|c}
& \!\!sensitivity\!\! & Taylor & $w^2$-rule & $z$-rules\\\hline
$\mathcal{X} = \mathbb{R}^d$      & \bf yes & \bf yes & \bf yes & no\\
$\mathcal{X} = \mathbb{R}^d_+,\mathcal{B}$ & \bf yes & \bf yes & no & \bf yes\\\hline
nearest root on $\mathcal{X}$\!\! & no & \bf yes & \bf yes & no\\\hline
\em conservative & \em no & \em no & \em yes & \em yes\\
\em positive & \em yes & \em \!yes$^\star$\! & \em yes & \em yes\\
consistent & no & no & \bf yes & \bf yes\\\hline
unique solution & \bf yes & \bf yes & \bf yes & \bf yes\\
fast computation & \bf yes & no & \bf yes & \bf yes
\end{tabular}
\end{center}
{\small $^{(\star)}$~e.g. using the continuously differentiable approximation of the detection function $\max(0,x) = \lim_{t \to \infty} t^{-1} \log(0.5+0.5\exp(t x))$.}

\begin{figure*}[t]
\centering \small
\begin{tabular}{c|c|c|c}
Sensitivity (rescaled) &
Taylor (nearest root) &
Deep Taylor ($w^2\!$-rule) &
Deep Taylor ($z^\mathcal{B}\!$-rule)\\
\includegraphics[width=0.225\linewidth]{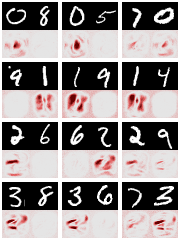}&
\includegraphics[width=0.225\linewidth]{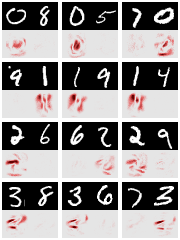}&
\includegraphics[width=0.225\linewidth]{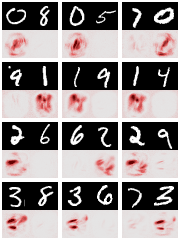}&
\includegraphics[width=0.225\linewidth]{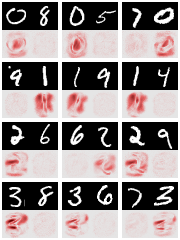}\\
\end{tabular}
\caption{\label{fig:shallow-heatmaps} Comparison of heatmaps produced by various decompositions. Each input image (pair of handwritten digits) is presented with its associated heatmap.}
\vskip 5mm
\begin{tabular}{c|c|c|c}
Sensitivity (rescaled) &
Taylor (nearest root) &
Deep Taylor ($w^2\!$-rule) &
Deep Taylor ($z^\mathcal{B}\!$-rule)\\
\includegraphics[width=0.225\linewidth]{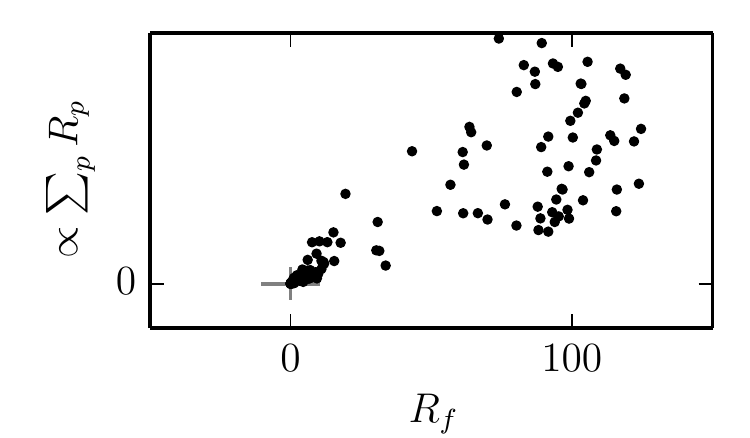}&
\includegraphics[width=0.225\linewidth]{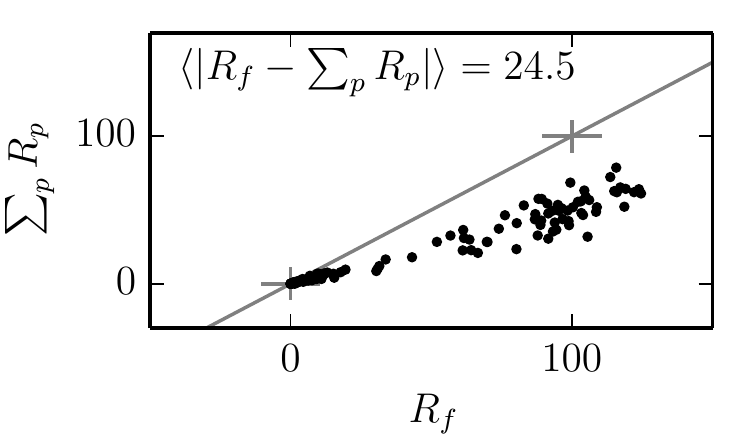}&
\includegraphics[width=0.225\linewidth]{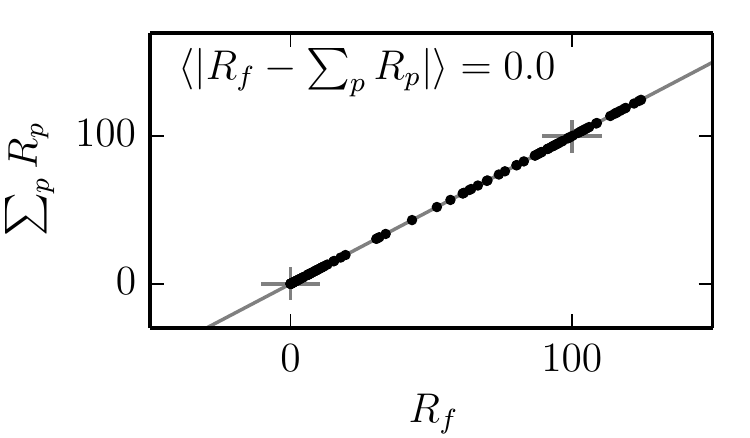}&
\includegraphics[width=0.225\linewidth]{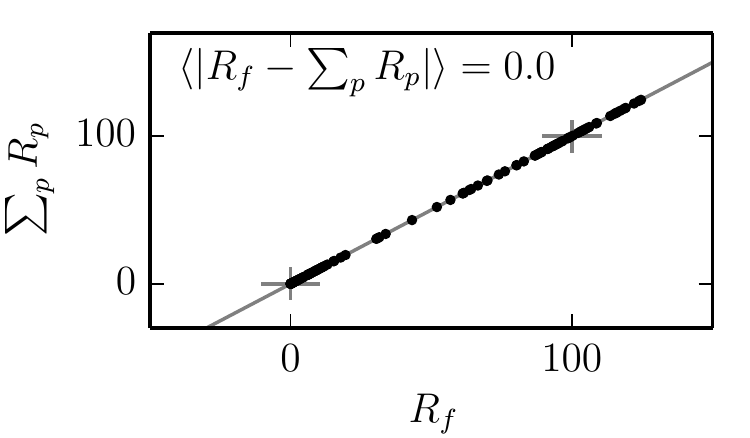}\\
\includegraphics[width=0.225\linewidth,trim=0 10 0 0]{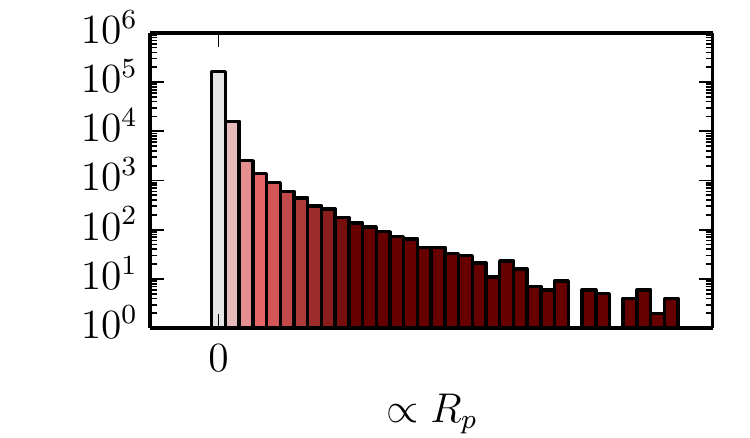}&
\includegraphics[width=0.225\linewidth,trim=0 10 0 0]{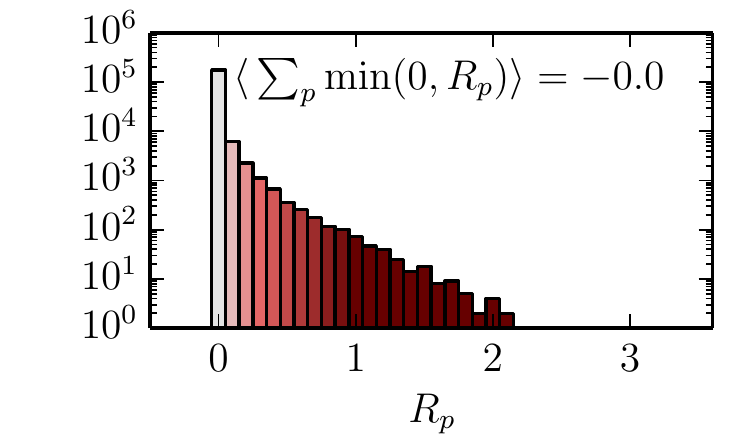}&
\includegraphics[width=0.225\linewidth,trim=0 10 0 0]{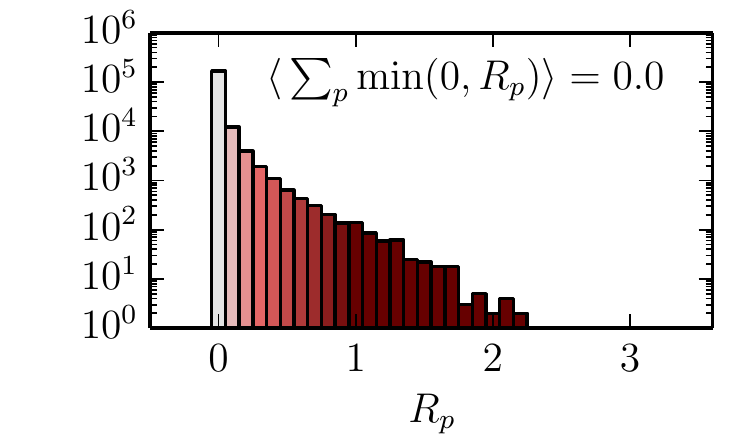}&
\includegraphics[width=0.225\linewidth,trim=0 10 0 0]{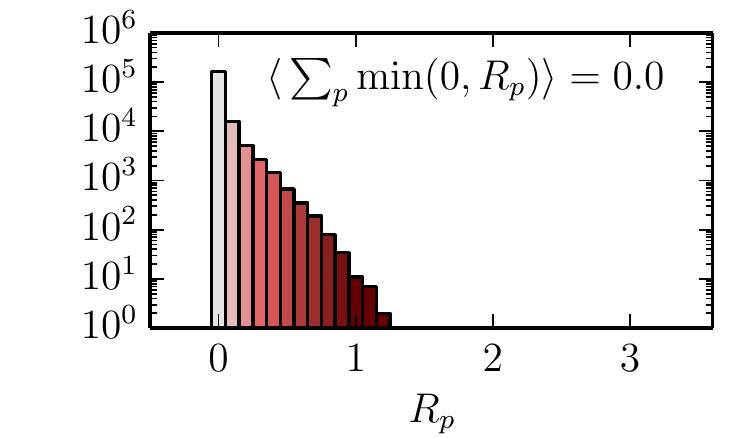}\\
\end{tabular}
\caption{\label{fig:shallow-scatter} Quantitative analysis of each decomposition technique. {\em Top:} Scatter plots showing for each data point the total output relevance (x-axis), and the sum of pixel-wise relevances (y-axis). {\em Bottom:} Histograms showing in log scale the number of times a particular value of pixel relevance occurs.}
\end{figure*}

\subsection{Experiment}

We now demonstrate empirically the properties of the heatmapping techniques introduced so far on the network of Figure \ref{fig:onelayer} trained to predict whether a MNIST handwritten digit of class 0--3 is present in the input image, next to a distractor digit of a different class 4--9. The neural network is trained to output $x_k = 0$ if there is no digit to detect in the image and $x_k = 100$ if there is one. We minimize the mean-square error between the true scores $\{0,100\}$, and the neural network output $x_k$. Treating the supervised task as a regression problem forces the network to assign approximately the same amount of relevance to all positive examples, and as little relevance as possible to the negative examples.

The input image is of size $28 \times 56$ pixels and is coded between $-0.5$ (black) and $+1.5$ (white). The neural network has $28 \times 56$ input neurons $\{x_i\}$, $400$ hidden neurons $\{x_j\}$, and one output $x_k$. Weights $\{w_{ij}\}$ are initialized using a normal distribution of mean $0$ and standard deviation $0.05$. Biases $\{b_j\}$ are initialized to zero and constrained to be negative or zero throughout training, in order to meet our specification of the one-layer network. The neural network is trained for $300000$ iterations of stochastic gradient descent with a minibatch of size $20$ and a small learning rate. Training data is extended with translated versions of MNIST digits. The root $\{\widetilde x_i\}$ for the nearest root Taylor method is chosen in our experiments to be the nearest point such that $f(\{\widetilde x_i\}) < 0.1 f(\{x_i\})$. The $z^\mathcal{B}\!$-rule is computed using as a lower- and upper-bounds $\forall_i: l_i = -0.5$ and $h_i = 1.5$.

Heatmaps are shown in Figure \ref{fig:shallow-heatmaps} for sensitivity analysis, nearest root Taylor decomposition, and deep Taylor decomposition with the $w^2$- and $z^\mathcal{B}\!$-rules. In all cases, we observe that the heatmapping procedure correctly assigns most of the relevance to pixels where the digit to detect is located. Sensitivity analysis produces unbalanced and incomplete heatmaps, with some input points reacting strongly, and others reacting weakly, and with a considerable amount of relevance associated to the border of the image, where there is no information. Nearest root Taylor produces selective heatmaps, that are still not fully complete. The heatmaps produced by deep Taylor decomposition with the $w^2$-rule, are similar to nearest root Taylor, but blurred, and not perfectly aligned with the data. The domain-aware $z^\mathcal{B}\!$-rule produces heatmaps that are still blurred, but that are complete and well-aligned with the data.

Figure \ref{fig:shallow-scatter} quantitatively evaluates heatmapping techniques of Figure \ref{fig:shallow-heatmaps}. The scatter plots compare the total output relevance with the sum of pixel-wise relevances. Each point in the scatter plot is a different data point drawn independently from the input distribution. These scatter plots test empirically for each heatmapping method whether it is conservative in the sense of Definition \ref{def:conservative}. In particular, if all points lie on the diagonal line of the scatter plot, then $\sum_p R_p = R_f$, and the heatmapping is conservative. The histograms just below test empirically whether the studied heatmapping methods satisfy positivity in the sense of Definition \ref{def:positive}, by counting the number of times (shown on a log-scale) pixel-wise contributions $R_p$ take a certain value. Red color in the histogram indicates positive relevance assignments, and blue color indicates negative relevance assignments. Therefore, an absence of blue bars in the histogram indicates that the heatmap is positive (the desired behavior). Overall, the scatter plots and the histograms produce a complete description of the degree of consistency of the heatmapping techniques in the sense of Definition \ref{def:consistent}.

Sensitivity analysis only measures a local effect and therefore does not conceptually redistribute relevance onto the input. However, we can still measure the relative strength of computed sensitivities between examples or pixels. The nearest root Taylor approach, although producing mostly positive heatmaps, dissipates a large fraction of the relevance. The deep Taylor decomposition on the other hand ensure full consistency, as theoretically predicted by Propositions \ref{prop:w2rule-consistent} and \ref{prop:zbrule-consistent}. The $z^\mathcal{B}$-rule spreads relevance onto more pixels than methods based on nearest root, as shown by the shorter tail of its relevance histogram.

\begin{figure*}
\centering
\includegraphics[width=0.95\textwidth]{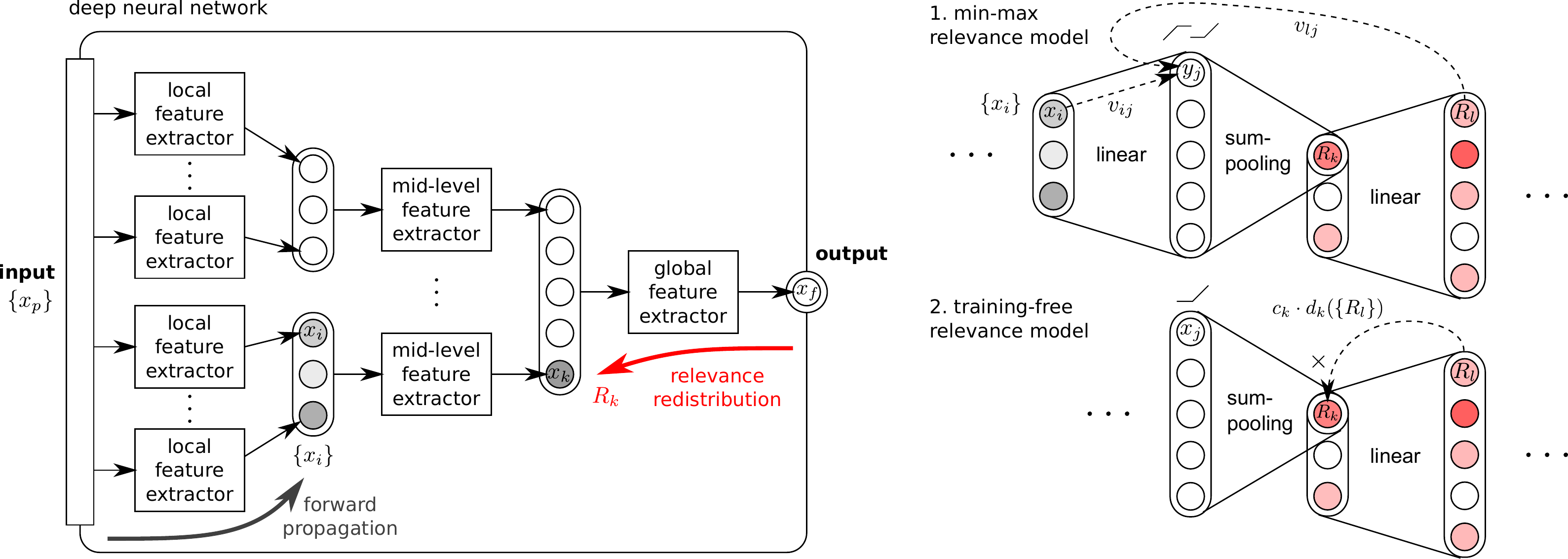}
\caption{Left: Example of a 3-layer deep network, composed of increasingly high-level feature extractors. Right: Diagram of the two proposed relevance models for redistributing relevance onto lower layers.}
\label{figure:hierarchy} 
\end{figure*}

\section{Application to Deep Networks}
\label{section:twolayers}

In order to represent efficiently complex hierarchical problems, one needs deeper architectures. These architectures are typically made of several layers of nonlinearity, where each layer extracts features at different scale. An example of deep architecture is shown in Figure \ref{figure:hierarchy} (left). In this example, the input is first processed by feature extractors localized in the pixel space. The resulting features are combined into more complex mid-level features that cover more pixels. Finally, these more complex features are combined in a final stage of nonlinear mapping, that produces a score determining whether the object to detect is present in the input image or not. A practical example of deep network with similar hierarchical architecture, and that is frequently used for image recognition tasks, is the convolutional neural network~\cite{lecun-89}.

In Section \ref{section:decomposition} and \ref{section:onelayer}, we have assumed the existence and knowledge of a functional mapping between the neuron activities at a given layer and relevances in the higher layer. However, in deep architectures, the mapping may be unknown (although it may still exist). In order to redistribute the relevance from the higher layers to the lower layers, one needs to make this mapping explicit. For this purpose, we introduce the concept of relevance model.

A relevance model is a function that maps a set of neuron activations at a given layer to the relevance of a neuron in a higher layer, and whose output can be redistributed onto its input variables, for the purpose of propagating relevance backwards in the network. For the deep network of Figure \ref{figure:hierarchy} (left), on can for example, try to predict $R_k$ from $\{x_i\}$, which then allows us to decompose the predicted relevance $R_k$ into lower-layer relevances $\{R_i\}$. For practical purposes, the relevance models we will consider borrow the structure of the one-layer network studied in Section \ref{section:onelayer}, and for which we have already derived a deep Taylor decomposition.

Upper-layer relevance is not only determined by input neuron activations of the considered layer, but also by high-level information (i.e. abstractions) that have been formed in the top layers of the network. These high-level abstractions are necessary to ensure a global cohesion between low-level parts of the heatmap.

\subsection{Min-Max Relevance Model}

We first consider a trainable relevance model of $R_k$. This relevance model is illustrated in Figure \ref{figure:hierarchy}-1 and is designed to incorporate both bottom-up and top-down information, in a way that the relevance can still be fully decomposed in terms of input neurons. It is defined as
\begin{align*}
y_j &= \max\big(0,\textstyle{\sum}_i x_i v_{ij} + a_j \big)\\
\widehat R_k &= \textstyle{\sum}_j  y_j.
\end{align*}
where $a_j = \min(0,\textstyle{\sum}_{l} R_{l} v_{lj} + d_j)$ is a negative bias that depends on upper-layer relevances, and where $\sum_l$ runs over the detection neurons of that upper-layer. This negative bias plays the role of an inhibitor, in particular, it prevents the activation of the detection unit $y_j$ of the relevance model in the case where no upper-level abstraction in $\{R_l\}$ matches the feature detected in $\{x_i\}$.

The parameters $\{v_{ij},v_{lj},d_j\}$ of the relevance model are learned by minimization of the mean square error objective
\begin{align*}
\min \big\langle (\widehat R_k - R_k )^2 \big\rangle,
\end{align*}
where $R_k$ is the true relevance, $\widehat R_k$ is the predicted relevance, and $\langle \cdot \rangle$ is the expectation with respect to the data distribution.

Because the relevance model has exactly the same structure as the one-layer neural network described in Section \ref{section:onelayer}, in particular, because $a_j$ is negative and only weakly dependent on the set of neurons $\{x_i\}$, one can apply the same set of rules for relevance propagation. That is, we compute
\begin{align}
\boxed{R_{j} = y_j}
\label{eq:rm-x}
\end{align}
for the pooling layer and
\begin{align}
\boxed{R_{i} = \sum_j \frac{q_{ij}}{\sum_{i'} q_{i'j}} R_{j}}
\label{eq:rm-q}
\end{align}
for the detection layer, where $q_{ij} = v_{ij}^2$, $q_{ij} = x_{i} v_{ij}^+$, or $q_{ij} = x_{i} v_{ij} - l_i v_{ij}^+ - h_i v_{ij}^-$ if choosing the $w^2$-, $z^+\!$-, $z^\mathcal{B}\!$-rules respectively. This set of equations used to backpropagate relevance from $R_k$ to $\{R_i\}$, is approximately conservative, with an approximation error that is determined by how much on average the output of the relevance model $\widehat R_k$ differs from the true relevance $R_k$.

\subsection{Training-Free Relevance Model}
\label{section:twolayers-mf}

A large deep neural network may have taken weeks or months to train, and we should be able to explain it without having to train a relevance model for each neuron. We consider the original feature extractor
\begin{align*}
x_j &= \max\big(0,\textstyle{\sum}_i x_i w_{ij} + b_j\big)\\
x_k &= \|\{x_j\}\|_p
\end{align*}
where the $L^p$-norm can represent a variety of pooling operations such as sum-pooling or max-pooling. Assuming that the upper-layer has been explained by the $z^+\!$-rule, and indexing by $l$ the detection neurons of that upper-layer, we can write the relevance $R_k$ as
\begin{align*}
R_k
&= \sum_l \frac{x_k w_{kl}^+}{\sum_{k'} x_{k'} w_{k'l}^+}  R_l\\
&= \big({\textstyle \sum}_j x_j\big) \cdot
\frac{\|\{x_j\}\|_p}{\|\{x_j\}\|_1} \cdot
\sum_l \frac{w_{kl}^+ R_l}{\sum_{k'} x_{k'} w_{k'l}^+}
\end{align*}
The first term is a linear pooling over detection units that has the same structure as the network of Section \ref{section:onelayer}. The second term is a positive $L^p/L^1$ pooling ratio, which is constant under any permutation of neurons $\{x_j\}$, or multiplication of these neurons by a scalar. The last term is a positive weighted sum of higher-level relevances, that measures the sensitivity of the neuron relevance to its activation. It is mainly determined by the relevance found in higher layers and can be viewed as a top-down contextualization term $d_k(\{R_l\})$. Thus, we rewrite the relevance as
$$
R_k = \big({\textstyle \sum}_j x_j\big) \cdot c_k \cdot d_k(\{R_l\})
$$
where the pooling ratio $c_k > 0$ and the top-down term $d_k(\{R_l\}) > 0$ are only weakly dependent on $\{x_j\}$ and are approximated as constant terms. This relevance model is illustrated in Figure \ref{figure:hierarchy}-2. Because the relevance model above has the same structure as the network of Section \ref{section:onelayer} (up to a constant factor), it is easy to derive its Taylor decomposition, in particular one can show that
\begin{align*}
\boxed{R_{j} = \frac{x_j}{\sum_{j'} x_{j'}} R_k}
\end{align*}
where relevance is redistributed in proportion to activations in the detection layer, and that
\begin{align*}
\boxed{R_{i} = \sum_j \frac{q_{ij}}{\sum_{i'} q_{i'j}} R_{j}}.
\end{align*}
where $q_{ij} = w_{ij}^2$, $q_{ij} = x_{i} w_{ij}^+$, or $q_{ij} = x_{i} w_{ij} - l_i w_{ij}^+ - h_i w_{ij}^-$ if choosing the $w^2$-, $z^+\!$-, $z^\mathcal{B}\!$-rules respectively. If choosing the $z^+\!$-rule for that layer again, the same training-free decomposition technique can be applied again to the layer below, and the process can be repeated until the input layer. Thus, when using the training-free relevance model, all layers of the network must be decomposed using the $z^+\!$-rule, except the first layer for which other rules can be applied such as the $w^2$-rule or the $z^\mathcal{B}\!$-rule.

The technical advantages and disadvantages of each heatmapping method are summarized in the table below:

\begin{center}
\begin{tabular}{l||c|c|c|c}
& \!\!sensitivity\!\! & \!Taylor\! & \!min-max\! & \!\parbox{11mm}{\centering training\\[-1mm]-free}\!\\\hline
consistent & no & no & \bf yes$^\dagger$ & \bf yes \\
unique solution & \bf yes & no$^\ddagger$ & no$^\ddagger$ & \bf yes\\
training-free & \bf yes & \bf yes & no & \bf yes\\
fast computation\!\! & \bf yes & no & \bf yes & \bf yes
\end{tabular}
\end{center}
{\small $^\dagger$ Conservative up to a fitting error between the redistributed relevance and the relevance model output. $^\ddagger$ Root finding and relevance model training are in the general case both nonconvex.}

\begin{figure*}[t]
\centering \small
\begin{tabular}{c|c|c|c}
Sensitivity (rescaled) &
Taylor (nearest root) &
Deep Taylor (min-max) &
Deep Taylor (training-free)\\
\includegraphics[width=0.225\linewidth]{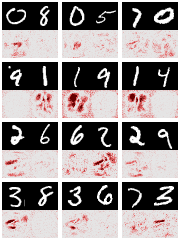} &
\includegraphics[width=0.225\linewidth]{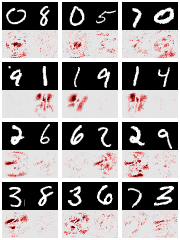} &
\includegraphics[width=0.225\linewidth]{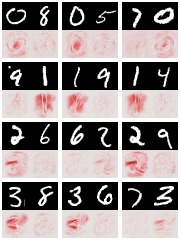} &
\includegraphics[width=0.225\linewidth]{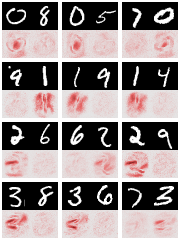}\\
\end{tabular}
\caption{\label{fig:deep-heatmaps} Comparison of heatmaps produced by various decompositions and relevance models. Each input image is presented with its associated heatmap.}
\vskip 5mm
\begin{tabular}{c|c|c|c}
Sensitivity (rescaled) &
Taylor (nearest root) &
Deep Taylor (min-max) &
Deep Taylor (training-free)\\
\includegraphics[width=0.225\linewidth]{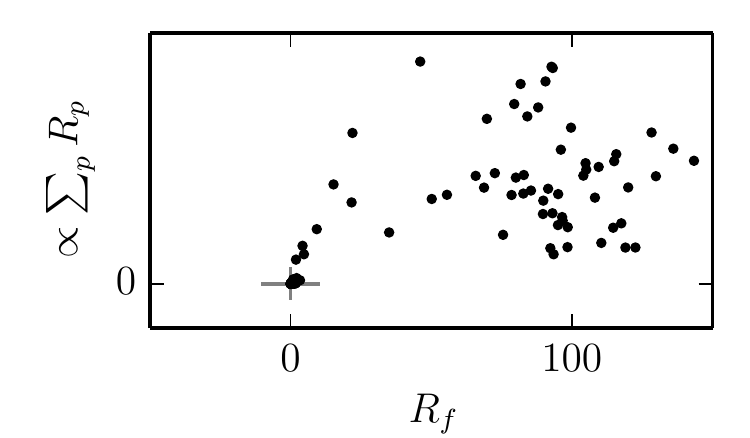}&
\includegraphics[width=0.225\linewidth]{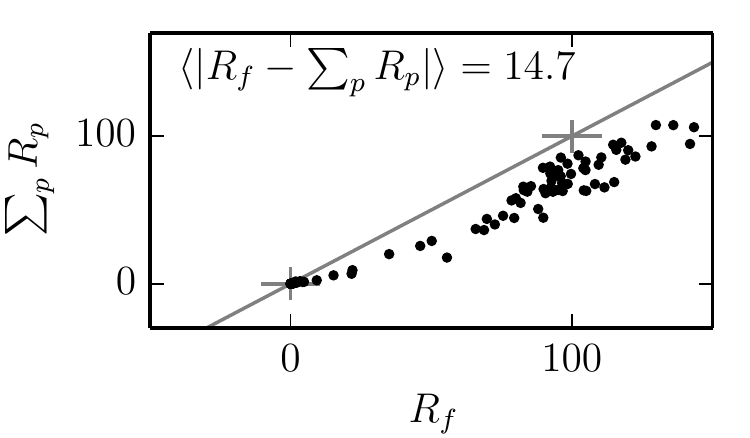} &
\includegraphics[width=0.225\linewidth]{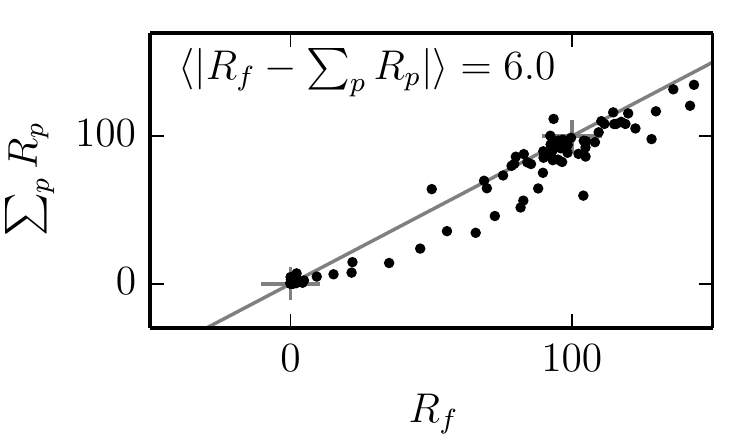} &
\includegraphics[width=0.225\linewidth]{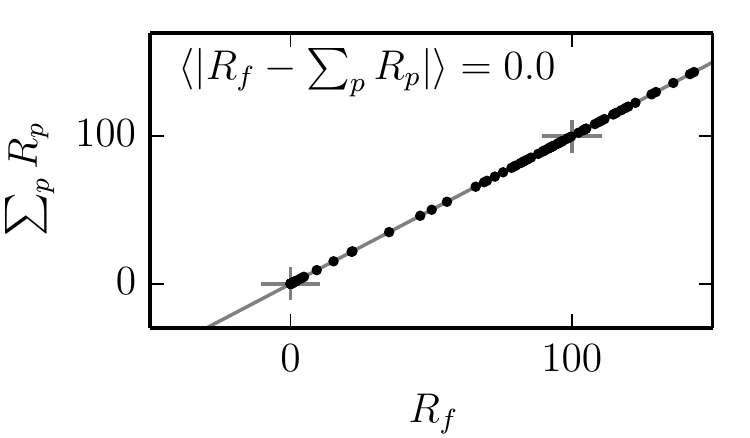}\\
\includegraphics[width=0.225\linewidth,trim=0 10 0 0]{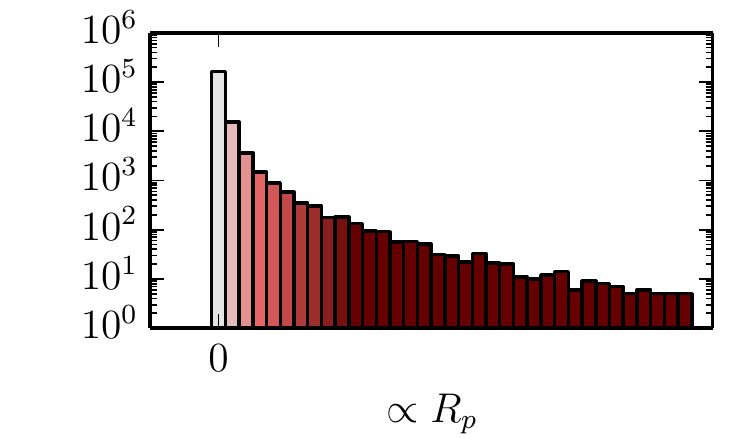}&
\includegraphics[width=0.225\linewidth,trim=0 10 0 0]{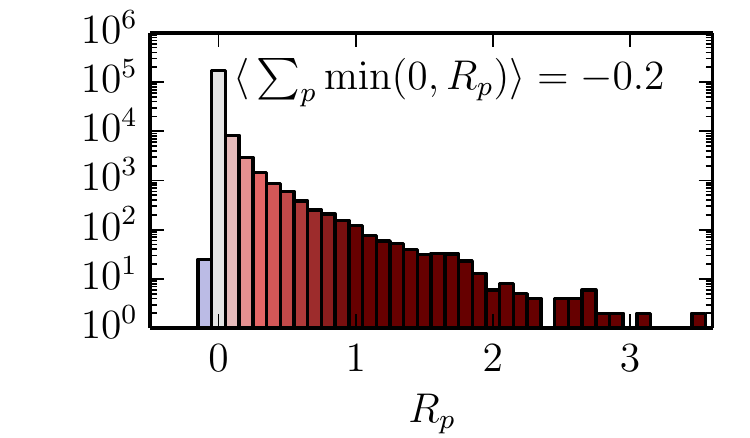}&
\includegraphics[width=0.225\linewidth,trim=0 10 0 0]{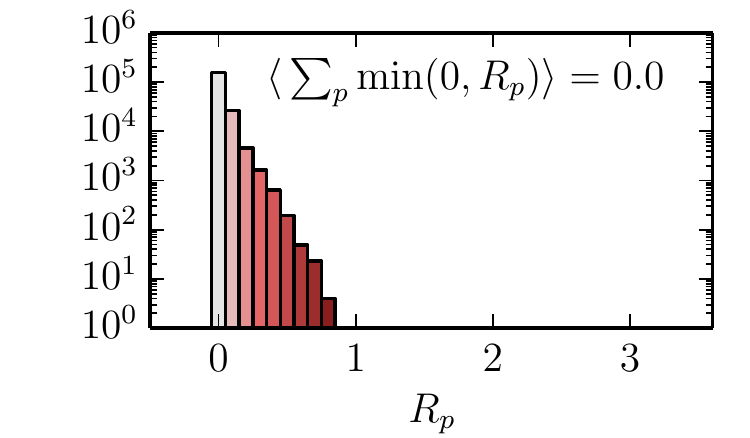}&
\includegraphics[width=0.225\linewidth,trim=0 10 0 0]{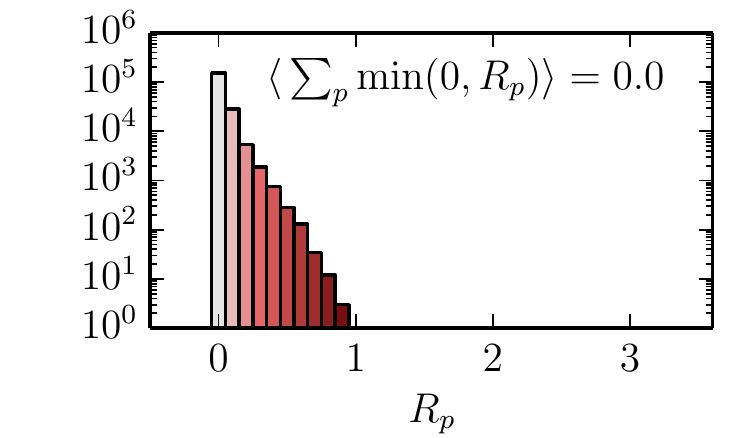}\\
\end{tabular}
\caption{\label{fig:deep-scatter} {\em Top:} Scatter plots showing for each type of decomposition and data points the predicted class score (x-axis), and the sum-of-relevance in the input layer (y-axis). {\em Bottom:} Histograms showing the number of times (on a log-scale) a particular pixel-wise relevance score occurs.}
\end{figure*}

\subsection{Experiment on MNIST}

We train a neural network with two layers of nonlinearity on the same MNIST problem as in Section \ref{section:onelayer}. The neural network is composed of a first detection-pooling layer with $400$ detection neurons sum-pooled into $100$ units (i.e. we sum-pool groups of 4 detection units). A second detection-pooling layer with $400$ detection neurons is applied to the resulting $100$-dimensional output of the previous layer, and activities are sum-pooled onto a single unit representing the deep network output. In addition, we learn a min-max relevance model for the first layer. The relevance model is trained to minimize the mean-square error between the relevance model output and the true relevance (obtained by application of the $z^+\!$-rule in the top layer). The deep network and the relevance models are trained using stochastic gradient descent with minibatch size $20$, for $300000$ iterations, and using a small learning rate.

Figure \ref{fig:deep-heatmaps} shows heatmaps obtained with sensitivity analysis, standard Taylor decomposition, and deep Taylor decomposition with different relevance models. We apply the $z^\mathcal{B}\!$-rule to backpropagate relevance of pooled features onto pixels. Sensitivity analysis and standard Taylor decomposition produce noisy and incomplete heatmaps. These two methods do not handle well the increased depth of the network. The min-max Taylor decomposition and the training-free Taylor decomposition produce relevance maps that are complete, and qualitatively similar to those obtained by deep Taylor decomposition of the shallow architecture in Section \ref{section:onelayer}. This demonstrates the high level of transparency of deep Taylor methods with respect to the choice of architecture. The heatmaps obtained by the trained min-max relevance model and by the training-free method are of similar quality.

Similar advantageous properties of the deep Taylor decomposition are observed quantitatively in the plots of Figure \ref{fig:deep-scatter}. The standard Taylor decomposition is positive, but dissipates relevance. The deep Taylor decomposition with the min-max relevance model produces near-conservative heatmaps, and the training-free deep Taylor decomposition produces heatmaps that are fully conservative. Both deep Taylor decomposition variants shown here also ensures positivity, due to the application of the $z^\mathcal{B}\!$- and $z^+\!$-rule in the respective layers.

\begin{figure*}
\centering \small
\begin{tabular}{cccc}
Image &
Sensitivity (CaffeNet) &
Deep Taylor (CaffeNet) &
Deep Taylor (GoogleNet)\\
\includegraphics[trim=0 35 0 35,clip,width=0.22\linewidth]{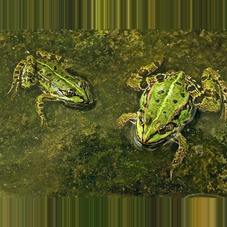}&
\includegraphics[trim=0 35 0 35,clip,width=0.22\linewidth]{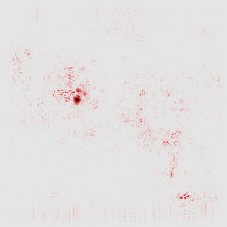}&
\includegraphics[trim=0 35 0 35,clip,width=0.22\linewidth]{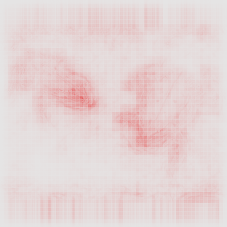}&
\includegraphics[trim=0 35 0 35,clip,width=0.22\linewidth]{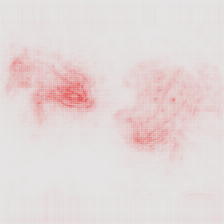}\\

\includegraphics[trim=0 35 0 35,clip,width=0.22\linewidth]{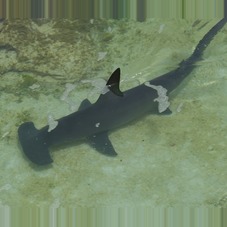}&
\includegraphics[trim=0 35 0 35,clip,width=0.22\linewidth]{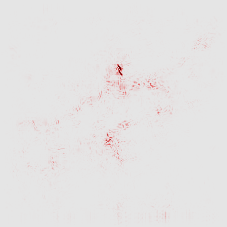}&
\includegraphics[trim=0 35 0 35,clip,width=0.22\linewidth]{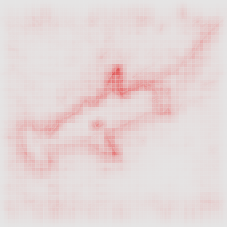}&
\includegraphics[trim=0 35 0 35,clip,width=0.22\linewidth]{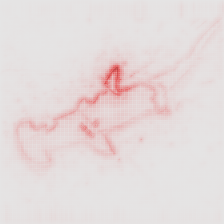}\\

\includegraphics[trim=0 25 0 45,clip,width=0.22\linewidth]{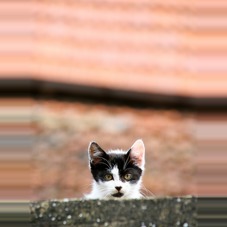}&
\includegraphics[trim=0 25 0 45,clip,width=0.22\linewidth]{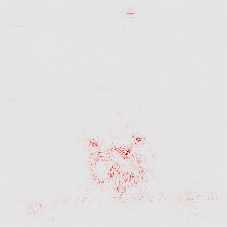}&
\includegraphics[trim=0 25 0 45,clip,width=0.22\linewidth]{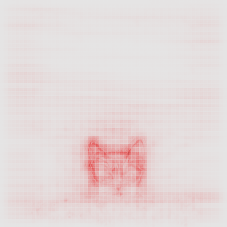}&
\includegraphics[trim=0 25 0 45,clip,width=0.22\linewidth]{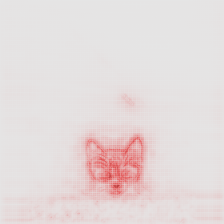}\\

\includegraphics[trim=0 35 0 35,clip,width=0.22\linewidth]{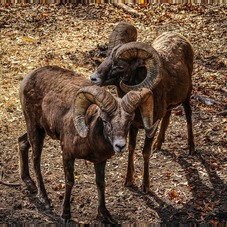}&
\includegraphics[trim=0 35 0 35,clip,width=0.22\linewidth]{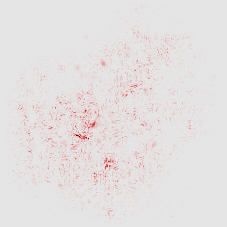}&
\includegraphics[trim=0 35 0 35,clip,width=0.22\linewidth]{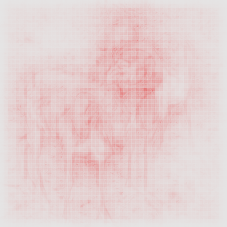}&
\includegraphics[trim=0 35 0 35,clip,width=0.22\linewidth]{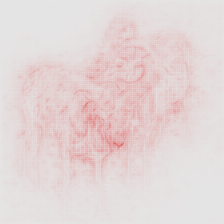}\\

\includegraphics[trim=0 35 0 35,clip,width=0.22\linewidth]{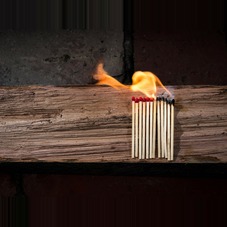}&
\includegraphics[trim=0 35 0 35,clip,width=0.22\linewidth]{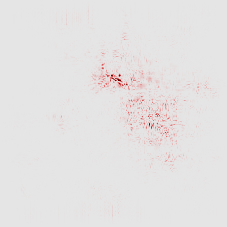}&
\includegraphics[trim=0 35 0 35,clip,width=0.22\linewidth]{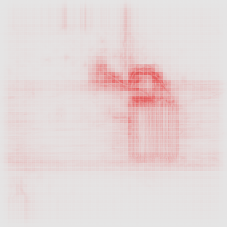}&
\includegraphics[trim=0 35 0 35,clip,width=0.22\linewidth]{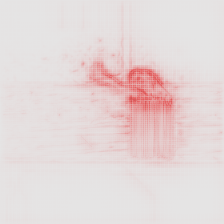}\\

\includegraphics[trim=0 35 0 35,clip,width=0.22\linewidth]{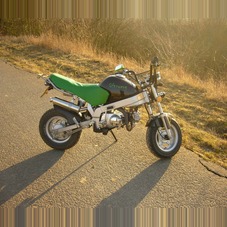}&
\includegraphics[trim=0 35 0 35,clip,width=0.22\linewidth]{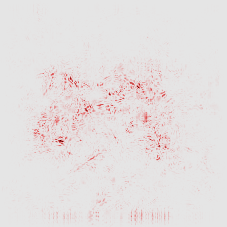}&
\includegraphics[trim=0 35 0 35,clip,width=0.22\linewidth]{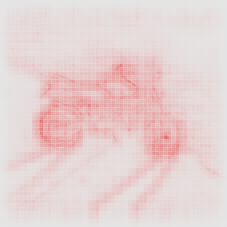}&
\includegraphics[trim=0 35 0 35,clip,width=0.22\linewidth]{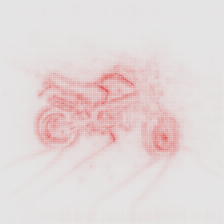}\\

\includegraphics[trim=0 45 0 25,clip,width=0.22\linewidth]{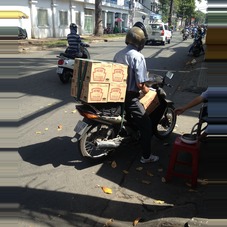}&
\includegraphics[trim=0 45 0 25,clip,width=0.22\linewidth]{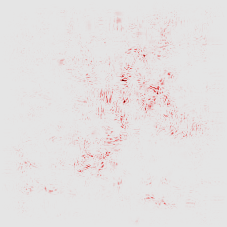}&
\includegraphics[trim=0 45 0 25,clip,width=0.22\linewidth]{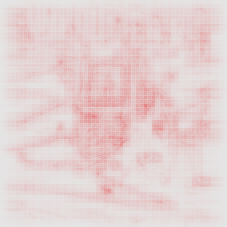}&
\includegraphics[trim=0 45 0 25,clip,width=0.22\linewidth]{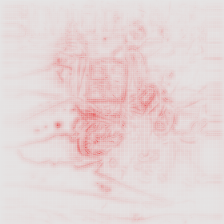}\\

\includegraphics[trim=0 45 0 25,clip,width=0.22\linewidth]{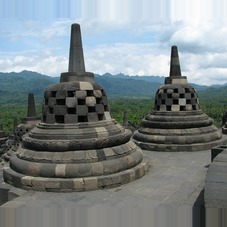}&
\includegraphics[trim=0 45 0 25,clip,width=0.22\linewidth]{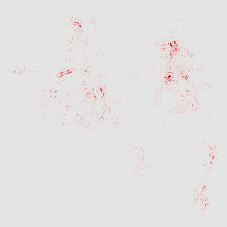}&
\includegraphics[trim=0 45 0 25,clip,width=0.22\linewidth]{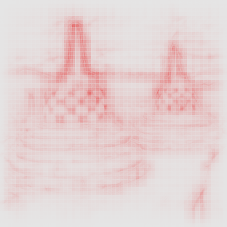}&
\includegraphics[trim=0 45 0 25,clip,width=0.22\linewidth]{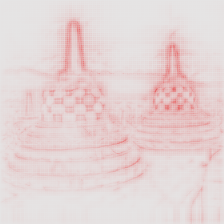}
\end{tabular}
\caption{Images of different ILSVRC classes (``frog'', ``shark'', ``cat'', ``sheep'', ``matchstick'', ``motorcycle'', ``scooter'', and ``stupa'') given as input to a deep network, and displayed next to the corresponding heatmaps. Heatmap scores are summed over all color channels.}
\label{fig:ilsvrc}
\end{figure*}

\subsection{Experiment on ILSVRC}

We now apply the fast training-free decomposition to explain decisions made by large neural networks (BVLC Reference CaffeNet \cite{Jia13caffe} and GoogleNet \cite{DBLP:journals/corr/SzegedyLJSRAEVR14}) trained on the dataset of the ImageNet large scale visual recognition challenges ILSVRC 2012 \cite{ilsvrc2012} and ILSVRC 2014 \cite{DBLP:journals/ijcv/RussakovskyDSKS15} respectively. For these models, standard Taylor decomposition methods with root finding are computationally too expensive. We keep the neural networks unchanged.

The training-free relevance propagation method is tested on a number of images from Pixabay.com and Wikimedia Commons. The $z^\mathcal{B}\!$-rule is applied to the first convolution layer. For all higher convolution and fully-connected layers, the $z^+\!$-rule is applied. Positive biases (that are not allowed in our deep Taylor framework), are treated as neurons, on which relevance can be redistributed (i.e. we add $\max(0,b_j)$ in the denominator of $z^\mathcal{B}\!$- and $z^\mathcal{+}\!$-rules). Normalization layers are bypassed in the relevance propagation pass. In order to visualize the heatmaps in the pixel space, we sum the relevances of the three color channels, leading to single-channel heatmaps, where the red color designates relevant regions.

Figure \ref{fig:ilsvrc} shows the resulting heatmaps for eight different images. Deep Taylor decomposition produces exhaustive heatmaps covering the whole object to detect. On the other hand, sensitivity analysis assigns most of the relevance to a few pixels. Deep Taylor heatmaps for the Caffenet and Googlenet have a high level of similarity, showing the transparency of the heatmapping method to the choice of deep network architecture. However, GoogleNet being more accurate, its corresponding heatmaps are also of better quality, with more heat associated to the truly relevant parts of the image. Heatmaps identify the dorsal fin of the shark, the head of the cat, the flame above the matchsticks, or the wheels of the motorbike. The heatmaps are able to detect two instances of the same object within a same image, for example, the two frogs and the two stupas. The heatmaps also ignore most of the distracting structure, such as the horizontal lines above the cat's head, the wood pattern behind the matches, or the grass behind the motorcycle. Sometimes, the object to detect is shown in a less stereotypical pose or can be confused with the background. For example, the sheeps in the top-right image are overlapping and superposed to a background of same color, and the scooter is difficult to separate from the complex and high contrast urban background. This confuses the network and the heatmapping procedure, and in that case, a significant amount of relevance is lost to the background.

\begin{figure}[t]
\centering \small
\includegraphics[width=1.0\linewidth]{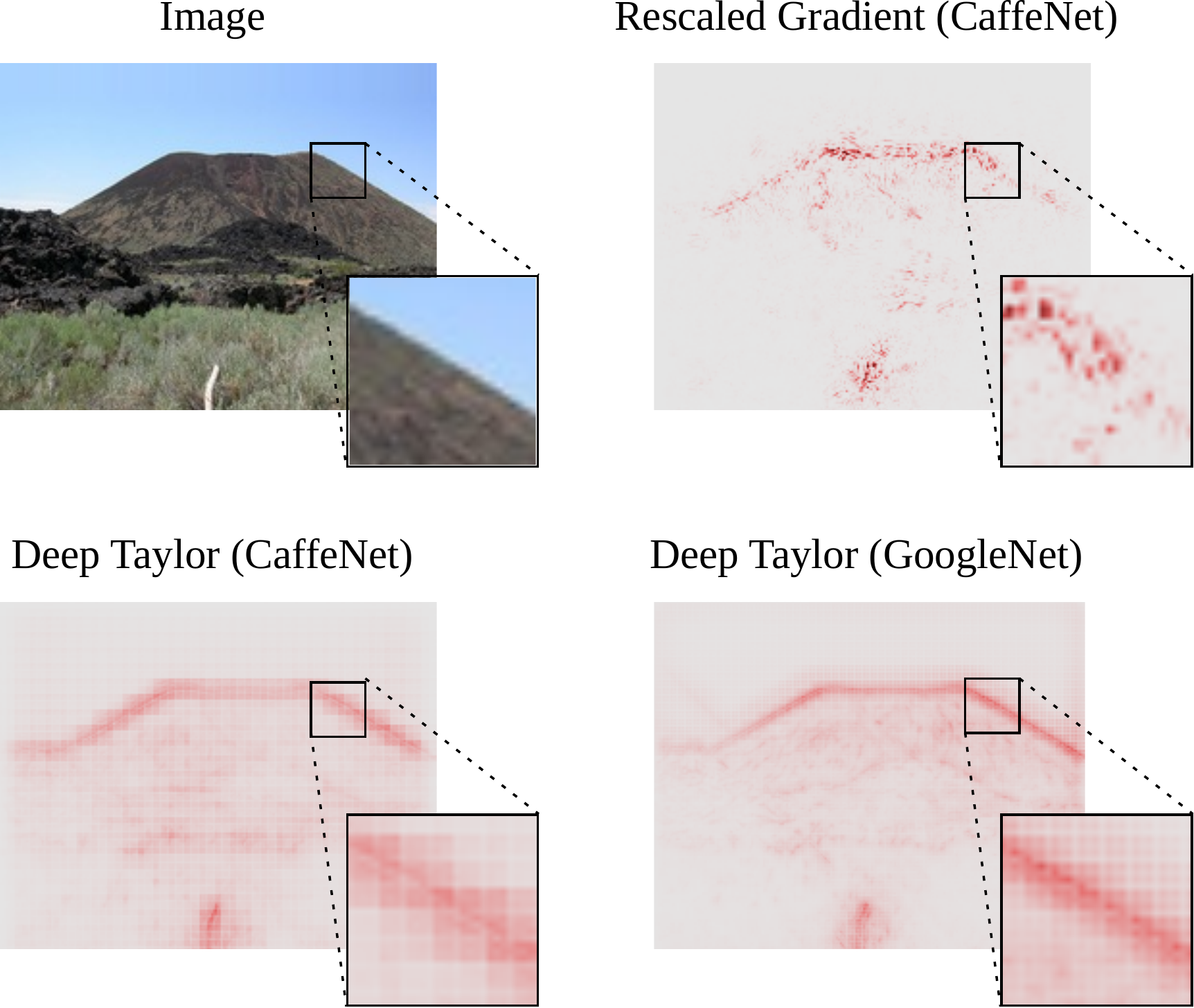}
\caption{Image with ILSVRC class ``volcano'', displayed next to its associated heatmaps and a zoom on a region of interest.}
\label{fig:zoom}
\end{figure}

Figure \ref{fig:zoom} studies the special case of an image of class ``volcano'', and a zoomed portion of it. On a global scale, the heatmapping method recognizes the characteristic outline of the volcano. On a local scale, the relevance is present on both sides of the edge of the volcano, which is consistent with the fact that the two sides of the edge are necessary to detect it. The zoomed portion of the image also reveals different stride sizes in the first convolution layer between CaffeNet (stride 4) and GoogleNet (stride 2). Therefore, our proposed heatmapping technique produces explanations that are interpretable both at a global and local scale in the pixel space.

\section{Conclusion}
\label{section:conclusion}

Nonlinear machine learning models have become standard tools in science and industry due to their excellent performance even for large, complex and high-dimensional problems. However, in practice it becomes more and more important to understand the underlying nonlinear model, i.e. to achieve transparency of {\em what} aspect of the input makes the model decide.

To achieve this, we have contributed by novel conceptual ideas to deconstruct nonlinear models. Specifically, we have proposed a novel relevance propagation approach based on deep Taylor decomposition, that is used to efficiently assess the importance of single pixels in image classification applications. Thus, we are now able to compute {\em heatmaps} that clearly and intuitively allow to better understand the role of input pixels when classifying an unseen data point.

In particular, we have shed light on theoretical connections between the Taylor decomposition of a function and rule-based relevance propagation techniques, showing a clear relationship between these two approaches for a particular class of neural networks. We have introduced the concept of relevance model as a mean to scale deep Taylor decomposition to neural networks with many layers. Our method is stable under different architectures and datasets, and does not require hyperparameter tuning.

We would like to stress, that we are free to use as a starting point of our framework either an own trained and carefully tuned neural network model or we may also download existing pre-trained deep network models (e.g. the Caffe Reference ImageNet Model \cite{Jia13caffe}) that have already been shown to achieve excellent performance on benchmarks. In both cases, our layer-wise relevance propagation concept can provide explanation. In other words our approach is orthogonal to the quest for enhanced results on benchmarks, in fact, we can use any benchmark winner and then enhance its transparency to the user.

\bibliographystyle{ieeetr}

\bibliography{paper}

\end{document}


\title{Explaining NonLinear Classification Decisions with Deep Taylor Decomposition\\[3mm]\sc \large (Supplementary Material)}

\author{Gr\'egoire Montavon,
        Sebastian Bach, 
        Alexander Binder,
        Wojciech Samek,
        and~Klaus-Robert M\"{u}ller}

\maketitle

\begin{abstract}
This supplement provides proofs, detailed derivations, pseudocode, and empirical comparisons with other relevance propagation techniques. 
\end{abstract}

\section{Derivations of Propagation Rules}
\label{section:rules-derivations}
In this section, we give the detailed derivations of propagation rules resulting from deep Taylor decomposition of the neural network of Section III of the paper. Each propagation rule corresponds to different choices of root point $\{\widetilde x_i\}^{(j)}$. For the class of networks considered here, the relevance of neurons in the detection layer is given by
\begin{align}
R_j = \max(0,{\textstyle\sum_i} x_i w_{ij} + b_j),
\label{eq:relmod}
\end{align}
where $b_j < 0$. All rules derived in this paper are based on the search for a root in a particular search direction $\{v_i\}^{(j)}$ in the input space associated to neuron $j$:
\begin{align}
\{\widetilde x_i\}^{(j)} = \{x_i\} + t \{v_i\}^{(j)}
\label{eq:search}
\end{align}
We need to consider two cases separately:
\begin{align*}
\mathcal{C}_1 &= \{j \colon {\textstyle\sum_i} x_i w_{ij} + b_j \leq 0\} = \{j \colon R_j = 0\}\\
\mathcal{C}_2 &= \{j \colon {\textstyle\sum_i} x_i w_{ij} + b_j > 0\} = \{j \colon R_j > 0\}
\end{align*}
In the first case ($j \in \mathcal{C}_1$), the data point itself is already the nearest root point of the function $R_j$. Therefore,
\begin{align}
x_i - \widetilde x_{i}^{(j)} = 0.
\label{eq:root0}
\end{align}
In the second case ($j \in \mathcal{C}_2$), the nearest root point along the defined search direction is given by the intersection of Equation \ref{eq:search} with the plane equation $\sum_i \widetilde x_{i}^{(j)} w_{ij} + b_j =  0$ to which the nearest root belong. In particular, resolving $t$ by injecting \eqref{eq:search} into that plane equation, we get
\begin{align}
x_{i} - \widetilde x_{i}^{(j)} =  \frac{\sum_i x_i w_{ij} + b_j}{\sum_i v_i^{(j)} w_{ij}} v_i^{(j)}
\label{eq:root}
\end{align}
Starting from the generic relevance propagation formula proposed in Section II we can derive a more specific formula that involve the search directions $\{v_{i}\}^{(j)}$:
\begin{align}
R_i
&= \sum_j \frac{\partial R_j}{\partial x_i} \Big|_{\{\widetilde x_{i}\}^{(j)}} \cdot (x_i - \widetilde x_{i}^{(j)}) \label{eq:relprop-original}\\
&= \sum_{j \in \mathcal{C}_1} \frac{\partial R_j}{\partial x_i} \cdot 0 + \sum_{j \in \mathcal{C}_2} w_{ij} \frac{\sum_i x_i w_{ij} + b_j}{\sum_i v_{i}^{(j)} w_{ij}} v_{i}^{(j)} \label{eq:relprop-split}\\
&= \sum_{j} \frac{v_{i}^{(j)} w_{ij}}{\sum_i v_{i}^{(j)} w_{ij}} R_j \label{eq:relprop}
\end{align}
From \eqref{eq:relprop-original} to \eqref{eq:relprop-split} we have considered the two listed cases separately, and injected their corresponding roots found in Equations \ref{eq:root0} and \ref{eq:root}. From \eqref{eq:relprop-split} to \eqref{eq:relprop}, we have used the fact that the relevance for the case $\mathcal{C}_1$ is always zero to recombine both terms.

\begin{figure*}
\centering \small
\includegraphics[width=1.0\linewidth]{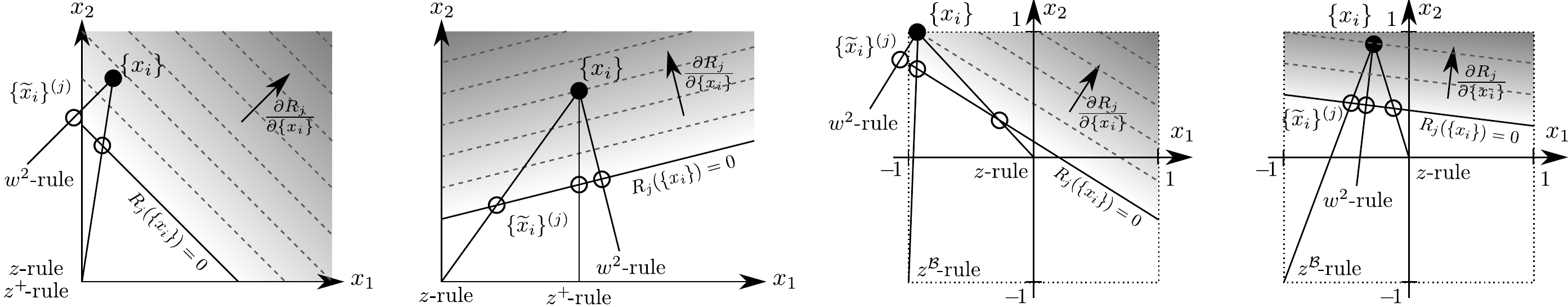}\vskip -2mm
\caption{Illustration of root points (empty circles) found for a given data point (full circle) for various propagation rules, relevance functions, and input domains. Here, for the $z^\mathcal{B}\!$-rule, we have used the bounding box $l_1 = -1$, $h_1 = 1$, $l_2 = -1$, $h_2 = 1$.}
\label{figure:rules-root}
\end{figure*}

The derivation of the various relevance propagation rules presented in this paper will always follow the same three steps:
\begin{enumerate}
\item Define for each neuron $j \in \mathcal{C}_2$ a line or segment in the input space starting from data point $\{x_i\}$ and with direction $\{v_{i}\}^{(j)}$.
\item Verify that the line or segment lies inside the input domain and includes at least one root of $R_j$.
\item Inject the search directions $\{v_{i}\}^{(j)}$ into Equation \ref{eq:relprop}, and obtain the relevance propagation rule as a result.
\end{enumerate}
An illustration of the search directions and root points selected by each rule for various relevance functions $R_j(\{x_i\})$ is given in Figure \ref{figure:rules-root}.

\subsection{w$^2$-Rule}

The $w^2$-rule is obtained by choosing the root of $R_j$ that is nearest to $\{x_i\}$ in $\mathbb{R}^d$. Such nearest root must be searched for on the line including the point $\{x_i\}$, and with direction corresponding to the gradient of $R_j$ (the $i$th component of this gradient is $w_{ij}$). Therefore, the components of the search vector are given by
$$
v_{i}^{(j)} = w_{ij}
$$
This line is included in the input domain $\mathbb{R}^d$, and always contains a root (the nearest of which is obtained by setting $t = -R_j / \sum_i w_{ij}^2$ in Equation \ref{eq:search}). Injecting the defined search direction $v_i$ into Equation \ref{eq:relprop}, we get
\begin{align*}
R_i &= \sum_j \frac{w_{ij}^2}{\sum_i w_{ij}^2} R_j.
\end{align*}

\subsection{z-Rule}

The $z$-rule (originally proposed by \cite{bach15}) is obtained by choosing the nearest root of $R_j$ on the segment $(\boldsymbol{0},\{x_i\})$. This segment is included in all domains considered in this paper ($\mathbb{R}^d, \mathbb{R}_+^d, \mathcal{B}$), provided that $\{x_i\}$ also belongs to these domains. This segment has a root at its first extremity, because $R_j(\boldsymbol{0}) = \max(0,\sum_i 0 \cdot w_{ij} + b_j) = \max(0,b_j) = 0$ since $b_j$ is negative by design. The direction of this segment on which we search for the nearest root corresponds to the data point itself:
$$
v_{i}^{(j)} = x_i.
$$
Injecting this search direction into Equation \ref{eq:relprop}, and defining the weighted activation $z_{ij} = x_i w_{ij}$, we get
\begin{align*}
R_i = \sum_j \frac{z_{ij}}{\sum_i z_{ij}} R_j.
\end{align*}

\subsection{z$^+\!$-Rule}

The $z^+\!$-rule is obtained by choosing the nearest root on the segment $(\{x_i 1_{w_{ij} < 0}\},\{x_i\})$. If $\{x_i\}$ is in $\mathbb{R}^d_+$, then, the segment is also in the domain $\mathbb{R}^d_+$. The relevance function has a root at the first extremity of the segment: \begin{align*}
R_j(\{x_i 1_{w_{ij} < 0}\})
&= \max(0,{\textstyle \sum_i} x_i 1_{w_{ij} < 0} w_{ij} + b_j)\\
&= \max(0,{\textstyle \sum_i} x_i w_{ij}^- + b_j) = 0,
\end{align*}
since $x_i \geq 0$ and $w_{ij}^- \leq 0$, and therefore $x_i w_{ij}^- \leq 0$, and since $b_j < 0$ by design. The direction of this segment on which we search for the nearest root is given by:
\begin{align*}
v_{i}^{(j)}
&= x_i - x_i 1_{w_{ij} < 0}\\
&= x_i 1_{w_{ij} \geq 0}.
\end{align*}
Injecting this search direction into Equation \ref{eq:relprop}, and defining $z_{ij}^+ = x_i w_{ij}^+$ with $w_{ij}^+ = 1_{w_{ij} \geq 0} w_{ij}$, we get
\begin{align*}
R_i = \frac{z_{ij}^+}{\sum_i z_{ij}^+} R_j.
\end{align*}

\subsection{z$^\mathcal{B}\!$-Rule}

The $z^\mathcal{B}\!$-rule is obtained by choosing the nearest root on the segment $(\{l_i 1_{w_{ij}>0} + h_i 1_{w_{ij}<0}\},\{x_i\})$. Provided that $\{x_i\}$ is in $\mathcal{B}$, the segment is also in $\mathcal{B}$. The relevance function has a root at the first extremity of the segment:
\begin{align*}
R_j(\{l_i & 1_{w_{ij}>0} + h_i 1_{w_{ij}<0}\})\\
&= \max(0,{\textstyle \sum_i} l_i 1_{w_{ij}>0} w_{ij} + h_i 1_{w_{ij}<0} w_{ij} + b_j)\\
&= \max(0,{\textstyle \sum_i} l_i w_{ij}^+ + h_i w_{ij}^- + b_j) = 0,
\end{align*}
because all summed terms are either negative or the product of a negative and positive value. The search direction for this choice of segment is given by
\begin{align*}
v_{i}^{(j)} &= x_i - l_i 1_{w_{ij}>0} - h_i 1_{w_{ij}<0}
\end{align*}
Injecting this search direction in to Equation \ref{eq:relprop}, we get
\begin{align*}
R_i = \sum_j \frac{z_{ij} -l_i w_{ij}^+ - h_i w_{ij}^-}{\sum_i z_{ij} -l_i w_{ij}^+ - h_i w_{ij}^-} R_j.
\end{align*}

\section{Algorithms for Propagation Rules}

We give here algorithms to implement the rules derived in Section \ref{section:rules-derivations} of the supplement. A useful property of these rules is that they can all be expressed in terms of matrix multiplications, thus, making them easily implementable with numerical libraries such as Matlab or Python/Numpy.

\subsection{w$^2$-Rule}

\noindent \begin{minipage}{1.0\linewidth}
\hrule \vskip 1mm
\begin{algorithmic}
\State \textbf{Input:}
\State \hskip 5mm Weight matrix $\texttt W = \{w_{ij}\}$
\State \hskip 5mm Upper-layer relevance vector $\texttt R = \{R_j\}$
\vskip 3mm
\State \textbf{Procedure:}
\State \hskip 5mm $\texttt V \leftarrow \texttt W \odot \texttt W$
\State \hskip 5mm $\texttt N \leftarrow \texttt V \oslash (\texttt{[1]} \cdot \texttt V)$
\State \hskip 5mm \textbf{return} $\texttt N \cdot \texttt R$
\end{algorithmic}
\hrule
\end{minipage} \vskip 2mm
\noindent where $\odot$ and $\oslash$ denote the element-wise multiplication and division respectively, and \texttt{[1]} is a matrix of ones. Note that for efficiency purposes, the squaring and normalization of the weight matrix can be performed once, and reused for many heatmaps computations.

\subsection{z-Rule}

\noindent \begin{minipage}{1.0\linewidth}
\hrule \vskip 1mm
\begin{algorithmic}
\State \textbf{Input:}
\State \hskip 5mm Weight matrix $\texttt W = \{w_{ij}\}$
\State \hskip 5mm Input activations $\texttt X = \{x_i\}$
\State \hskip 5mm Upper-layer relevance vector $\texttt R = \{R_j\}$
\vskip 3mm
\State \textbf{Procedure:}
\State \hskip 5mm $\texttt Z \leftarrow \texttt W^\top \texttt X$
\State \hskip 5mm \textbf{return} $\texttt X \odot (\texttt W \cdot (\texttt R \oslash \texttt Z))$
\end{algorithmic}
\hrule
\end{minipage} \vskip 2mm
\noindent where $\odot$ and $\oslash$ denote the element-wise multiplication and division respectively, and where the variable \texttt{Z} is the sum of weighted activations for each upper-layer neuron.

\subsection{z$^+\!$-Rule}

\noindent \begin{minipage}{1.0\linewidth}
\hrule \vskip 1mm
\begin{algorithmic}
\State \textbf{Input:}
\State \hskip 5mm Weight matrix $\texttt W = \{w_{ij}\}$
\State \hskip 5mm Input activations $\texttt X = \{x_i\}$
\State \hskip 5mm Upper-layer relevance vector $\texttt R = \{R_j\}$
\vskip 3mm
\State \textbf{Procedure:}
\State \hskip 5mm $\texttt V \leftarrow \texttt W^+$
\State \hskip 5mm $\texttt Z \leftarrow \texttt V^\top \texttt X$
\State \hskip 5mm \textbf{return} $\texttt X \odot (\texttt V \cdot (\texttt R \oslash \texttt Z))$
\end{algorithmic}
\hrule
\end{minipage} \vskip 2mm
\noindent where $\odot$ and $\oslash$ denote the element-wise multiplication and division respectively, and where the operation $(\cdot)^+$ keeps the positive part of the input matrix. For efficiency, like for the $w^2$-rule, the matrix \texttt{V} can be precomputed and reused for multiple heatmaps computations.

\subsection{z$^\mathcal{B}$-Rule}

\noindent \begin{minipage}{1.0\linewidth}
\hrule \vskip 1mm
\begin{algorithmic}
\State \textbf{Input:}
\State \hskip 5mm Weight matrix $\texttt W = \{w_{ij}\}$
\State \hskip 5mm Input activations $\texttt X = \{x_i\}$
\State \hskip 5mm Upper-layer relevance vector $\texttt R = \{R_j\}$
\State \hskip 5mm Lower-bound $\texttt L = \{l_i\}$
\State \hskip 5mm Upper-bound $\texttt H = \{h_i\}$
\vskip 3mm
\State \textbf{Procedure:}
\State \hskip 5mm $\texttt U \leftarrow \texttt W^-$
\State \hskip 5mm $\texttt V \leftarrow \texttt W^+$
\State \hskip 5mm $\texttt N \leftarrow \texttt R \oslash (\texttt W^\top \texttt X - \texttt V^\top \texttt L  - \texttt U^\top \texttt H)$
\State \hskip 5mm \textbf{return} $\texttt X \odot (\texttt W \cdot \texttt N) - \texttt L \odot (\texttt V \cdot \texttt N) - \texttt H \odot (\texttt U \cdot \texttt N)$
\end{algorithmic}
\hrule
\end{minipage} \vskip 2mm
\noindent where $\odot$ and $\oslash$ denote the element-wise multiplication and division respectively, and where the operations $(\cdot)^+,(\cdot)^-$ keep the positive part and the negative part of the input matrix respectively. For efficiency, like for the previous rules, the matrices $\texttt U$ and $\texttt V$ can be precomputed and reused for multiple heatmaps computations.

\section{Proofs of Propositions}

\begin{definition}
\label{def:conservative} A heatmapping $\R(\x)$ is \underline{conservative} if the sum of assigned relevances in the pixel space corresponds to the total relevance detected by the model, that is
\begin{align*}
\forall \x:~f(\x) = \sum_p R_p(\x).
\end{align*}
\end{definition}

\begin{definition}
\label{def:positive}
A heatmapping $\R(\x)$ is p\!\!\underline{\,\,ositive} if all values forming the heatmap are greater or equal to zero, that is:
\begin{align*}
\forall \x,p:~R_p(\x) \geq 0
\end{align*}
\end{definition}

\begin{definition}
\label{def:consistent}
A heatmapping $\R(\x)$ is \underline{consistent} if it is conservative \underline{and} positive. That is, it is consistent if it complies with Definitions \ref{def:conservative} and \ref{def:positive}.
\end{definition}

\begin{proposition}
For all $g \in \mathcal{G}$, the deep Taylor decomposition with the $w^2$-rule is consistent in the sense of Definition \ref{def:consistent}.
\label{prop:w2rule-consistent}
\end{proposition}

\begin{proof}
We first show that the heatmapping is conservative:
\begin{align*}
\sum_i R_i &= \sum_i \Big(\sum_j \frac{w_{ij}^2}{\sum_{i} w_{ij}^2} R_j\Big)\\
 &= \sum_j \frac{\sum_i w_{ij}^2}{\sum_{i} w_{ij}^2} R_j = \sum_j R_j = \sum_j x_j = f(x).
\end{align*}
where we have assumed the weights to be never exactly zero. Then, we show that the heatmapping is positive:
\begin{align*}
R_i
= \sum_j \frac{w_{ij}^2}{\sum_{i} w_{ij}^2} R_j 
= \sum_j
\underbrace{w_{ij}^2}_{> 0} \cdot
\underbrace{\frac{1}{\sum_{i} w_{ij}^2}}_{>0} \cdot
\underbrace{R_j}_{\geq 0} \geq 0.
\end{align*}
Therefore, because the heatmapping is both conservative and positive, it is also consistent.

For the case where $\sum_i w_{ij}^2 = 0$, it implies that $w_{ij} = 0$ for all $i$ and therefore $z_{ij} = 0$ for all $i$ too. Because $b_j \leq 0$, then $R_j = x_j = 0$ (there is no relevance to redistribute to the lower layer).
\end{proof}

\begin{proposition}
For all $g \in \mathcal{G}$ and data points $\{x_i\} \in \mathbb{R}_+^d$, the deep Taylor decomposition with the $z^+\!$-rule is consistent in the sense of Definition \ref{def:consistent}.
\label{prop:zprule-consistent}
\end{proposition}
\begin{proof}
The proof is the same as for Proposition \ref{prop:w2rule-consistent} for the case where $\sum_i z_{ij}^+ > 0$. We simply replace $w_{ij}^2$ by $z_{ij}^+$ in the proof.

For the case where $\sum_i z_{ij}^+ = 0$, it implies that $z_{ij} \leq 0$ for all $i$. Because $b_j \leq 0$, then $R_j = x_j = 0$ (there is no relevance to redistribute to the lower layer).
\end{proof}

\begin{proposition}
For all $g \in \mathcal{G}$ and data points $\{x_i\} \in \mathcal{B}$, the deep Taylor decomposition with the $z^\mathcal{B}\!$-rule is consistent in the sense of Definition \ref{def:consistent}.
\label{prop:zbrule-consistent}
\end{proposition}
\begin{proof}
We first show that the numerator of the $z^\mathcal{B}\!$-rule $q_{ij} = z_{ij} - l_i w_{ij}^+ - h_i w_{ij}^-$ is greater or equal than zero for $\{x_i\} \in \mathcal{B}$:
\begin{align*}
q_{ij} &= z_{ij} - l_i w_{ij}^+ - h_i w_{ij}^-\\
&= x_i w_{ij} - l_i w_{ij}^+ - h_i w_{ij}^-\\
&= x_i (w_{ij}^- + w_{ij}^+) - l_i w_{ij}^+ - h_i w_{ij}^-\\
&=
\underbrace{(x_i-h_i)}_{\leq 0} \cdot
\underbrace{w_{ij}^-}_{\leq 0} +
\underbrace{(x_i-l_i)}_{\geq 0} \cdot
\underbrace{w_{ij}^+}_{\geq 0} \geq 0
\end{align*}
Then, the proof is the same as for Proposition \ref{prop:w2rule-consistent} for the case where $\sum_i q_{ij} > 0$. We simply replace $w_{ij}^2$ by $q_{ij}$ in the proof. For the case where $\sum_i q_{ij} = 0$, we will show that the contributions $z_{ij}$ of the inputs to the detection neurons cannot be positive, and that there is therefore no relevance that needs to be redistributed. The equality $\sum_i q_{ij} = 0$ implies that $\forall_i:q_{ij} = 0$, which can be satisfied by one of the four sets of conditions:

\paragraph{$x_i = h_i$ and $x_i = l_i$} In that case $x_i = 0$ (because $l_i \leq 0$ and $h_i \geq 0$), and therefore $z_{ij} = 0 \cdot w_{ij} = 0$.

\paragraph{$x_i = h_i$ and $w_{ij}^+ = 0$} In that case, $z_{ij} = h_i w_{ij}$, and because $h_i \geq 0$, then $z_{ij} \leq 0$.

\paragraph{$w_{ij}^- = 0$ and $x_i = l_i$} In that case, $z_{ij} = l_i w_{ij}$, and because $l_i \leq 0$, then $z_{ij} \leq 0$.

\paragraph{$w_{ij}^- = 0$ and $w_{ij}^+ = 0$} In that case, $w_{ij}$ = 0, and therefore, $z_{ij} = x_i \cdot 0 = 0$.

Therefore, inputs are in all cases prevented from contributing positively to the neuron $x_j$. In particular, the total contribution is given by $z_j = \sum_i z_{ij} \leq 0$. Because $b_j \leq 0$, then $R_j = x_j = 0$ (there is no relevance to redistribute to the lower layer).
\end{proof}

\section{Empirical Comparison with LRP}

In this section, we compare heatmaps produced by the rules based on deep Taylor decomposition, and the layer-wise relevance propagation (LRP) rules proposed by \cite{bach15}. The last rules include in particular, the $\alpha\beta$-rule:
\begin{align*}
R_i &= \sum_j \Big( \alpha \frac{z_{ij}^+}{\sum_i z_{ij}^+ + b_j^+} - \beta \frac{z_{ij}^-}{\sum_i z_{ij}^- + b_j^-} \Big) R_j,
\end{align*}
where $\alpha-\beta = 1$, and the $\epsilon$-stabilized rule:
\begin{align*}
R_i &= \sum_j \frac{z_{ij}}{s(\sum_i z_{ij} + b_j)} R_j,
\end{align*}
where $s(t) = t+\epsilon (1_{t\geq 0} - 1_{t < 0})$ is a stabilizing function whose output is never zero. The respective hyperparameters $\alpha$ and $\epsilon$ of these rules are typically selected such that the produced heatmaps have the desired quality.

\begin{figure*}[t]
\centering \small
\begin{tabular}{c|ccc}
\bf Deep Taylor & \multicolumn{3}{c}{{\bf LRP} \cite{bach15}}\\
($z^\mathcal{B}\!$-rule) & (alphabeta, $\alpha=1,\beta=0$) & (alphabeta, $\alpha=2,\beta=1$) & (stabilized, $\epsilon=10$)\\
\includegraphics[width=0.225\linewidth]{results/heatmap-SNN-zb.png}&
\includegraphics[width=0.225\linewidth]{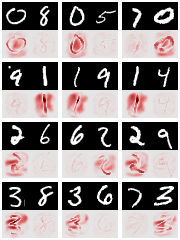} &
\includegraphics[width=0.225\linewidth]{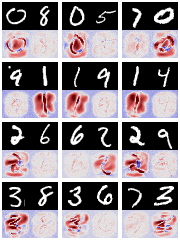}&
\includegraphics[width=0.225\linewidth]{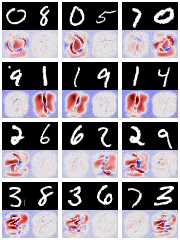}\\
\includegraphics[width=0.225\linewidth]{results/scatter-SNN-zb.pdf}&
\includegraphics[width=0.225\linewidth]{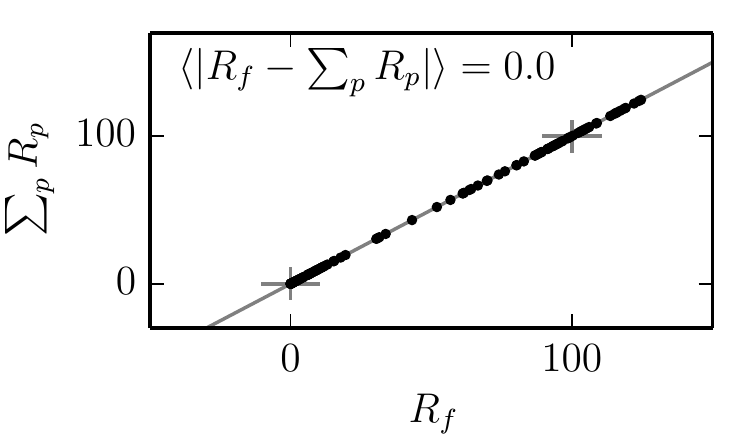}&
\includegraphics[width=0.225\linewidth]{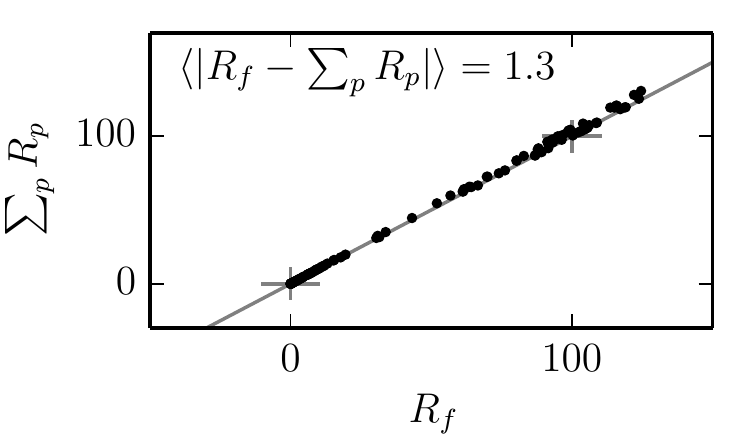}&
\includegraphics[width=0.225\linewidth]{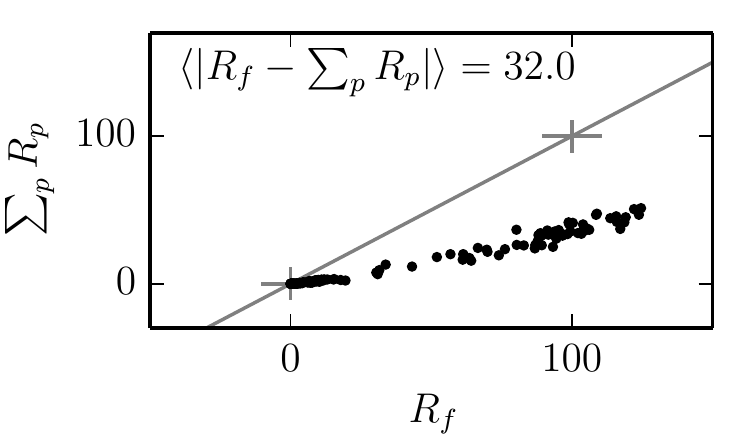}\\
\includegraphics[width=0.225\linewidth,trim=0 10 0 0]{results/histogram-SNN-zb.pdf}&
\includegraphics[width=0.225\linewidth,trim=0 10 0 0]{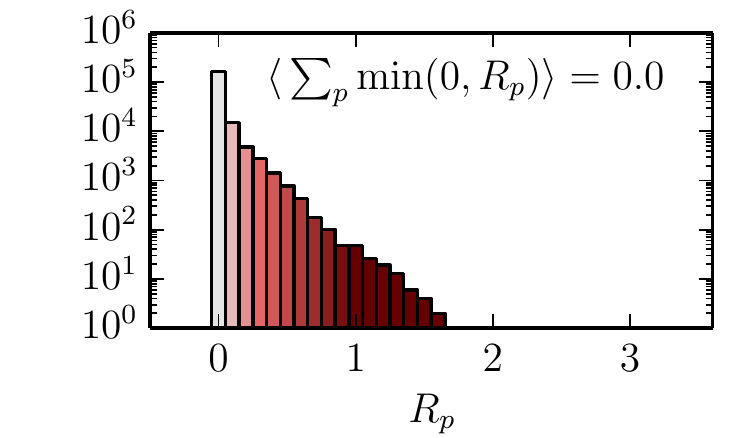}&
\includegraphics[width=0.225\linewidth,trim=0 10 0 0]{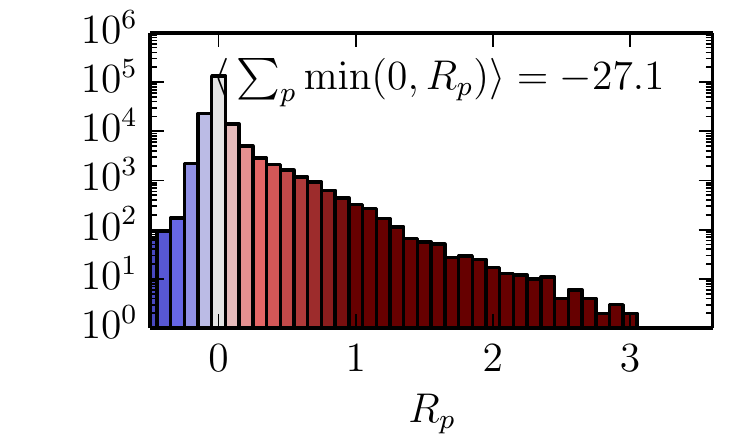}&
\includegraphics[width=0.225\linewidth,trim=0 10 0 0]{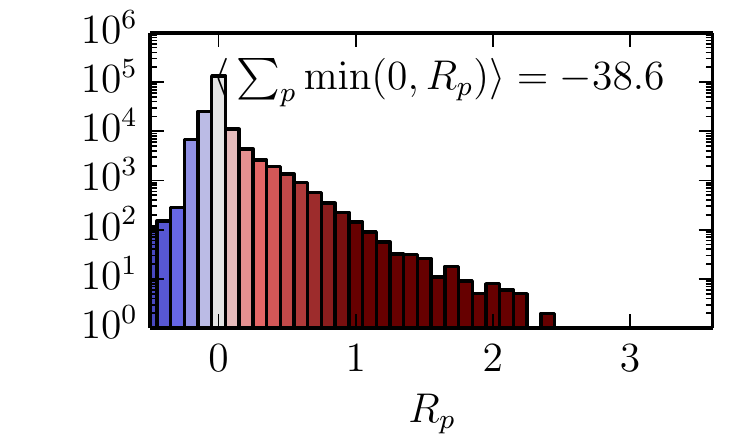}\\
\end{tabular}
\caption{Heatmaps and their properties produced by various heatmapping techniques applied on the one-layer network of Section III of the manuscript.}
\label{fig:onelayer}
\vskip 3mm
\centering \small
\begin{tabular}{c|ccc}
\bf Deep Taylor & \multicolumn{3}{c}{{\bf LRP} \cite{bach15}}\\
 (min-max, $z$-rules) & (alphabeta, $\alpha=1,\beta=0$) & (alphabeta, $\alpha=2,\beta=1$) & (stabilized, $\epsilon=3$)\\
\includegraphics[width=0.225\linewidth]{results/heatmap-MRM.png}&
\includegraphics[width=0.225\linewidth]{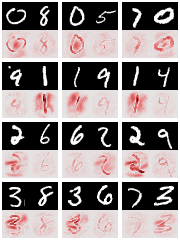} &
\includegraphics[width=0.225\linewidth]{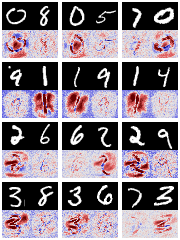}&
\includegraphics[width=0.225\linewidth]{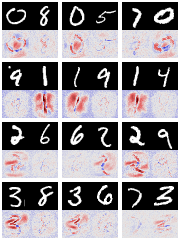} \\
\includegraphics[width=0.225\linewidth]{results/scatter-MRM.pdf}&
\includegraphics[width=0.225\linewidth]{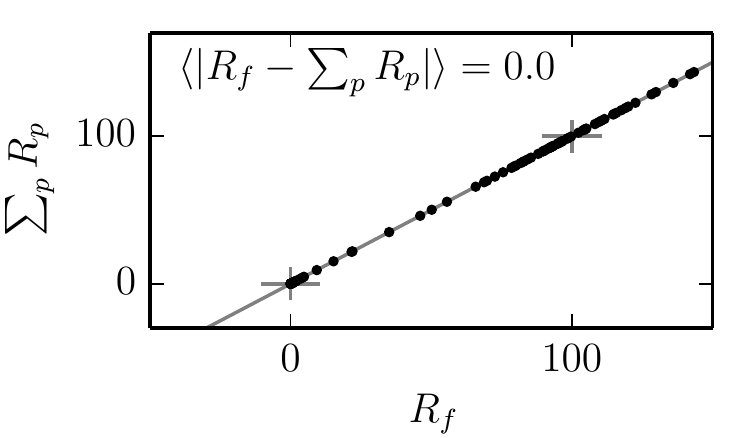}&
\includegraphics[width=0.225\linewidth]{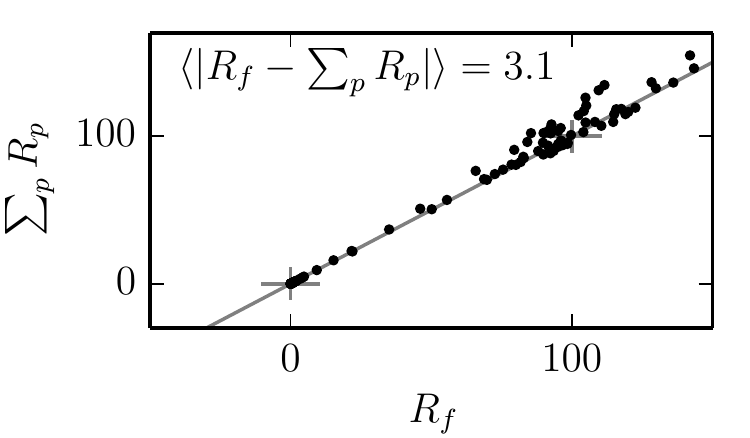}&
\includegraphics[width=0.225\linewidth]{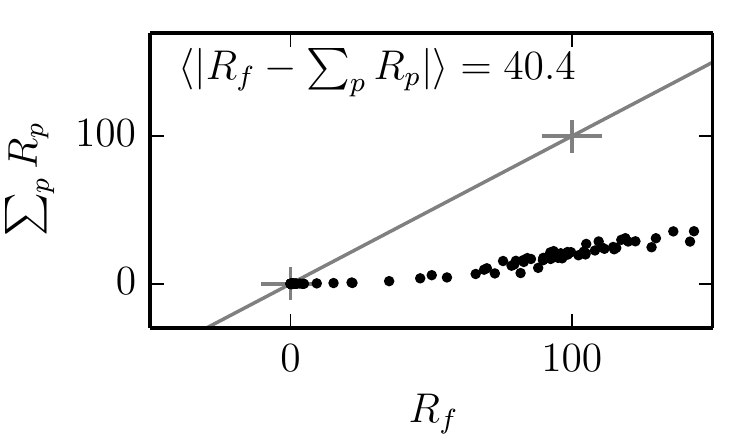}\\
\includegraphics[width=0.225\linewidth,trim=0 10 0 0]{results/histogram-MRM.pdf}&
\includegraphics[width=0.225\linewidth,trim=0 10 0 0]{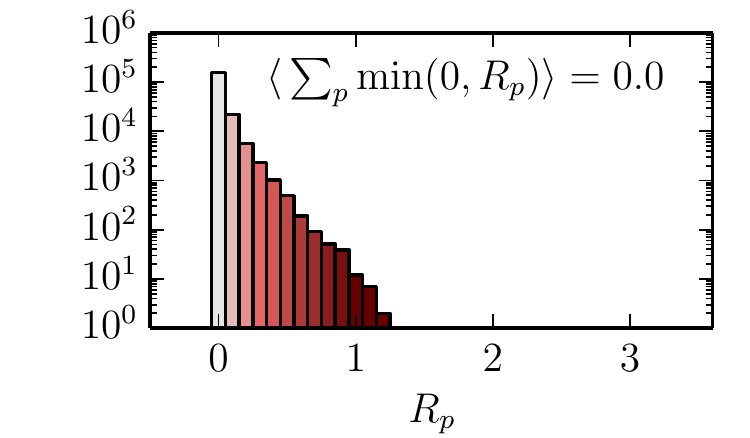}&
\includegraphics[width=0.225\linewidth,trim=0 10 0 0]{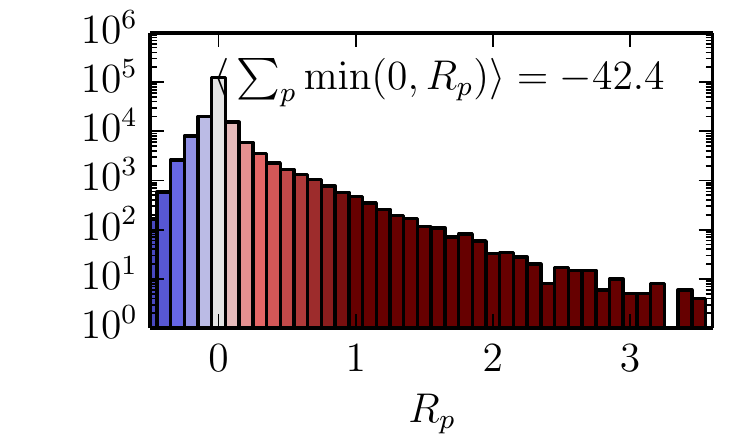}&
\includegraphics[width=0.225\linewidth,trim=0 10 0 0]{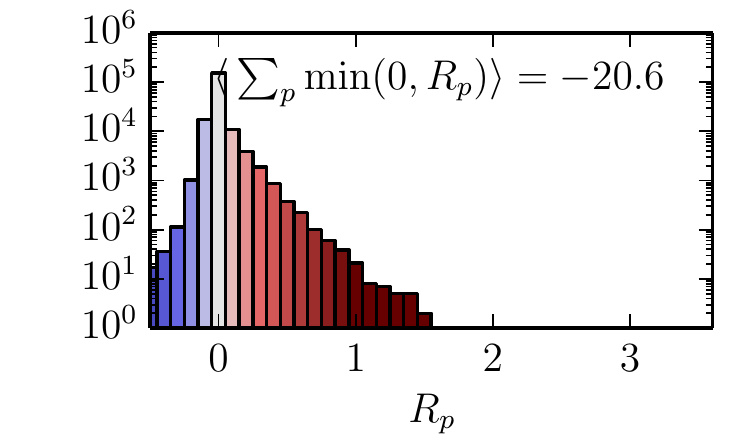}\\
\end{tabular}
\caption{Heatmaps and their properties produced by various heatmapping techniques applied on the two-layers network of Section IV of the manuscript.}
\label{fig:twolayers}
\end{figure*}

\begin{figure*}[t]
\centering \small
\begin{tabular}{cc|cc}
\multicolumn{2}{c}{\bf Deep Taylor} & \multicolumn{2}{c}{{\bf LRP} \cite{bach15} ($\alpha=2,\beta=1$)} \\
CaffeNet & GoogleNet & CaffeNet & GoogleNet\\
\includegraphics[trim=0 35 0 35,clip,width=0.22\linewidth]{images/bvlc_deeptaylor/frogs2-animal-178429_1280_hm.png}&
\includegraphics[trim=0 35 0 35,clip,width=0.22\linewidth]{images/gnet_deeptaylor/frogs2-animal-178429_1280_hm.png}&
\includegraphics[trim=0 35 0 35,clip,width=0.22\linewidth]{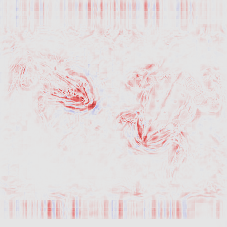}&
\includegraphics[trim=0 35 0 35,clip,width=0.22\linewidth]{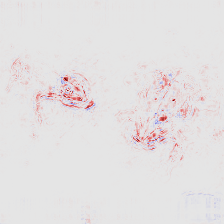}\\

\includegraphics[trim=0 35 0 35,clip,width=0.22\linewidth]{images/bvlc_deeptaylor/hammerhead-shark-298238_1280_hm.png}&
\includegraphics[trim=0 35 0 35,clip,width=0.22\linewidth]{images/gnet_deeptaylor/hammerhead-shark-298238_1280_hm.png}&
\includegraphics[trim=0 35 0 35,clip,width=0.22\linewidth]{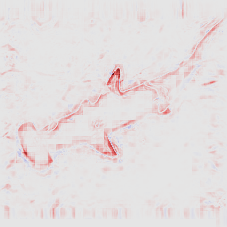}&
\includegraphics[trim=0 35 0 35,clip,width=0.22\linewidth]{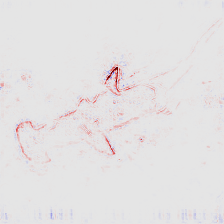}\\

\includegraphics[trim=0 25 0 45,clip,width=0.22\linewidth]{images/bvlc_deeptaylor/kitten-288549_1280_hm.png}&
\includegraphics[trim=0 25 0 45,clip,width=0.22\linewidth]{images/gnet_deeptaylor/kitten-288549_1280_hm.png}&
\includegraphics[trim=0 25 0 45,clip,width=0.22\linewidth]{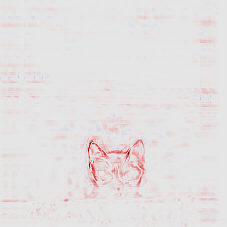}&
\includegraphics[trim=0 25 0 45,clip,width=0.22\linewidth]{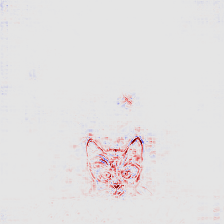}\\

\includegraphics[trim=0 35 0 35,clip,width=0.22\linewidth]{images/bvlc_deeptaylor/bighorn-sheep-204693_1280_hm.png}&
\includegraphics[trim=0 35 0 35,clip,width=0.22\linewidth]{images/gnet_deeptaylor/bighorn-sheep-204693_1280_hm.png}&
\includegraphics[trim=0 35 0 35,clip,width=0.22\linewidth]{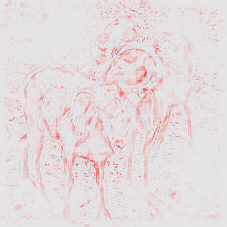}&
\includegraphics[trim=0 35 0 35,clip,width=0.22\linewidth]{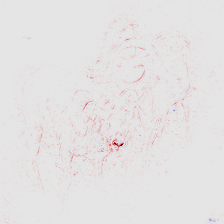}\\

\includegraphics[trim=0 35 0 35,clip,width=0.22\linewidth]{images/bvlc_deeptaylor/matches-171732_1280_hm.png}&
\includegraphics[trim=0 35 0 35,clip,width=0.22\linewidth]{images/gnet_deeptaylor/matches-171732_1280_hm.png}&
\includegraphics[trim=0 35 0 35,clip,width=0.22\linewidth]{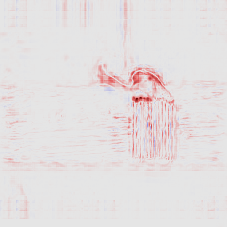}&
\includegraphics[trim=0 35 0 35,clip,width=0.22\linewidth]{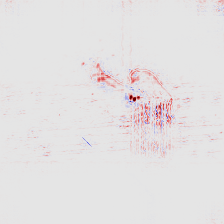}\\

\includegraphics[trim=0 35 0 35,clip,width=0.22\linewidth]{images/bvlc_deeptaylor/motorcycle-345001_1280_hm.png}&
\includegraphics[trim=0 35 0 35,clip,width=0.22\linewidth]{images/gnet_deeptaylor/motorcycle-345001_1280_hm.png}&
\includegraphics[trim=0 35 0 35,clip,width=0.22\linewidth]{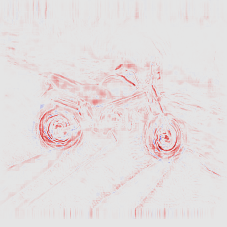}&
\includegraphics[trim=0 35 0 35,clip,width=0.22\linewidth]{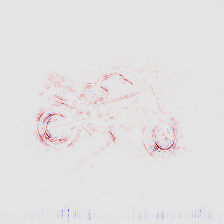}\\

\includegraphics[trim=0 45 0 25,clip,width=0.22\linewidth]{images/bvlc_deeptaylor/scooter_vietnam-265807_1280_hm.png}&
\includegraphics[trim=0 45 0 25,clip,width=0.22\linewidth]{images/gnet_deeptaylor/scooter_vietnam-265807_1280_hm.png}&
\includegraphics[trim=0 45 0 25,clip,width=0.22\linewidth]{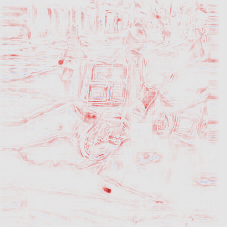}&
\includegraphics[trim=0 45 0 25,clip,width=0.22\linewidth]{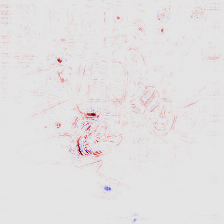}\\

\includegraphics[trim=0 45 0 25,clip,width=0.22\linewidth]{images/bvlc_deeptaylor/stupa-83774_1280_hm.png}&
\includegraphics[trim=0 45 0 25,clip,width=0.22\linewidth]{images/gnet_deeptaylor/stupa-83774_1280_hm.png}&
\includegraphics[trim=0 45 0 25,clip,width=0.22\linewidth]{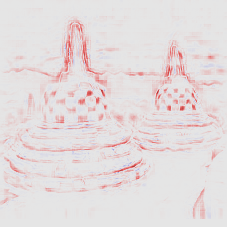}&
\includegraphics[trim=0 45 0 25,clip,width=0.22\linewidth]{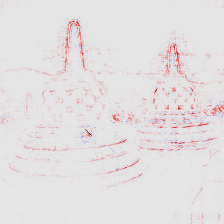}\\
\end{tabular}
\caption{Heatmaps produced by deep Taylor decomposition and LRP when applied to the predictions of the CaffeNet and GoogleNet networks.}
\label{fig:ilsvrc}
\end{figure*}

Figure \ref{fig:onelayer} and  \ref{fig:twolayers} compare heatmaps obtained by applying deep Taylor and LRP to the one-layer and two-layer networks considered in this paper for the MNIST problem. In the two-layer case, deep Taylor uses the min-max relevance model. LRP applies the same rules in both layers. Figure \ref{fig:ilsvrc} compares heatmaps obtained by deep Taylor and LRP on the BVLC CaffeNet and the GoogleNet. For deep Taylor, we use the training-free relevance model. Normalization layers are ignored in the backward pass.

It can be observed that the quality of the deep Taylor heatmaps is less influenced by the choice of model and dataset than LRP with a fixed set of parameters. Deep Taylor heatmaps look similar in all cases. LRP also produces high-quality heatmaps, but the best parameters differ in each setting. For example, the parameters $\alpha=2,\beta=1$ perform well for the CaffeNet, but tend to produce too sparse heatmaps for the GoogleNet, or to produce a large amount of negative relevance on the MNIST dataset. Various parameters of LRP produce various artefacts such as the presence of residual relevance on the irrelevant MNIST digit, or the presence of negative relevance in the black areas of the MNIST images. On the other hand, LRP-based heatmaps are sharper than Taylor-based heatmaps and less subject to the stride artefact that arises with convolutional neural networks. They also tend to assign very little evidence to irrelevant parts of the image. Future work will seek to identify the reason for the superiority of LRP on these particular aspects, and investigate whether the deep Taylor decomposition method and its underlying principles can be refined to incorporate these desirable properties of a heatmap while retaining its stability.

\bibliographystyle{ieeetr}

\bibliography{paper}